\documentclass[journal]{IEEEtran}
\usepackage{cite}
\ifCLASSINFOpdf
\else
\fi
\usepackage[pdftex]{graphicx}

\usepackage{amsmath}
\usepackage{amssymb}
\usepackage{bm}

\DeclareMathOperator*{\argmin}{arg\,min}

\usepackage{algorithm,algpseudocode}
\usepackage{multirow}

\usepackage{subfigure}

\begin{document}
	\title{Temporal Sparse Adversarial Attack on Sequence-based Gait Recognition}
	
	\author{Ziwen He, \IEEEmembership{Student Member, IEEE}, Wei Wang, \IEEEmembership{Member, IEEE}, Jing Dong, \IEEEmembership{Senior Member, IEEE},\\ and Tieniu Tan, \IEEEmembership{Fellow, IEEE}% <-this % stops a space
		\thanks{Z. He is with the Center for Research on Intelligent Perception and Computing, Institute of Automation, Chinese Academy of Sciences, Beijing 100190, China, also with the University of Chinese Academy of Sciences, Beijing 100049, China (e-mail: ziwen.he@cripac.ia.ac.cn).}% <-this % stops a space
		\thanks{W. Wang and T. Tan are with the Center for Research on Intelligent Perception and Computing, Institute of Automation, Chinese Academy of Sciences, Beijing 100190, China (e-mail: wwang@nlpr.ia.ac.cn; tnt@nlpr.ia.ac.cn).}% <-this % stops a space
		\thanks{J. Dong is with the Center for Research on Intelligent Perception and Computing, Institute of Automation, Chinese Academy of Sciences, Beijing 100190, China, and also with the State Key Laboratory of Information Security, Institute of Information Engineering, Chinese Academy of Sciences, Beijing 100093, China (e-mail: jdong@nlpr.ia.ac.cn).}% <-this % stops a space
	}

	\maketitle
	
	\begin{abstract}
		Gait recognition is widely used in social security applications due to its advantages in long-distance human identification. Recently, sequence-based methods have achieved high accuracy by learning abundant temporal and spatial information. However, their robustness under adversarial attacks has not been clearly explored. In this paper, we demonstrate that the state-of-the-art gait recognition model is vulnerable to such attacks. To this end, we propose a novel temporal sparse adversarial attack method. Different from previous additive noise models which add perturbations on original samples, we employ a generative adversarial network based architecture to semantically generate adversarial high-quality gait silhouettes or video frames. Moreover, by sparsely substituting or inserting a few adversarial gait silhouettes, the proposed method ensures its imperceptibility and achieves a high attack success rate. The experimental results show that if only one-fortieth of the frames are attacked, the accuracy of the target model drops dramatically.
	\end{abstract}
	

	% \fi
	\IEEEpeerreviewmaketitle

	\section{Introduction}
	
	\label{sec:intro}
	%gait recognition
	\IEEEPARstart{G}{ait} recognition is designed to automatically identify people according to their way of walking. Compared to traditional biometric information such as fingerprints or irises, gaits can be obtained at long distances without the cooperation of subjects. As a result, gait recognition is widely applied in remote visual surveillance solutions. In recent years, numerous gait recognition methods~\cite{Wu2016ACS,Takemura2019OnIA,He2019MultiTaskGF,Wolf2016MultiviewGR,Liao2017PoseBasedTN} have been proposed; they have achieved a high recognition accuracy. However, the security of gait recognition algorithms against malicious attacks has not been thoroughly studied. Limited work~\cite{Jia2019AttackingGR,DBLP1,DBLP2,DBLP3} has studied the robustness of gait recognition to spoofing attacks, which aim to gain illegitimate access to the target gait recognition models.
	
\begin{figure}
		\centering
		\includegraphics[width=1.0\linewidth]{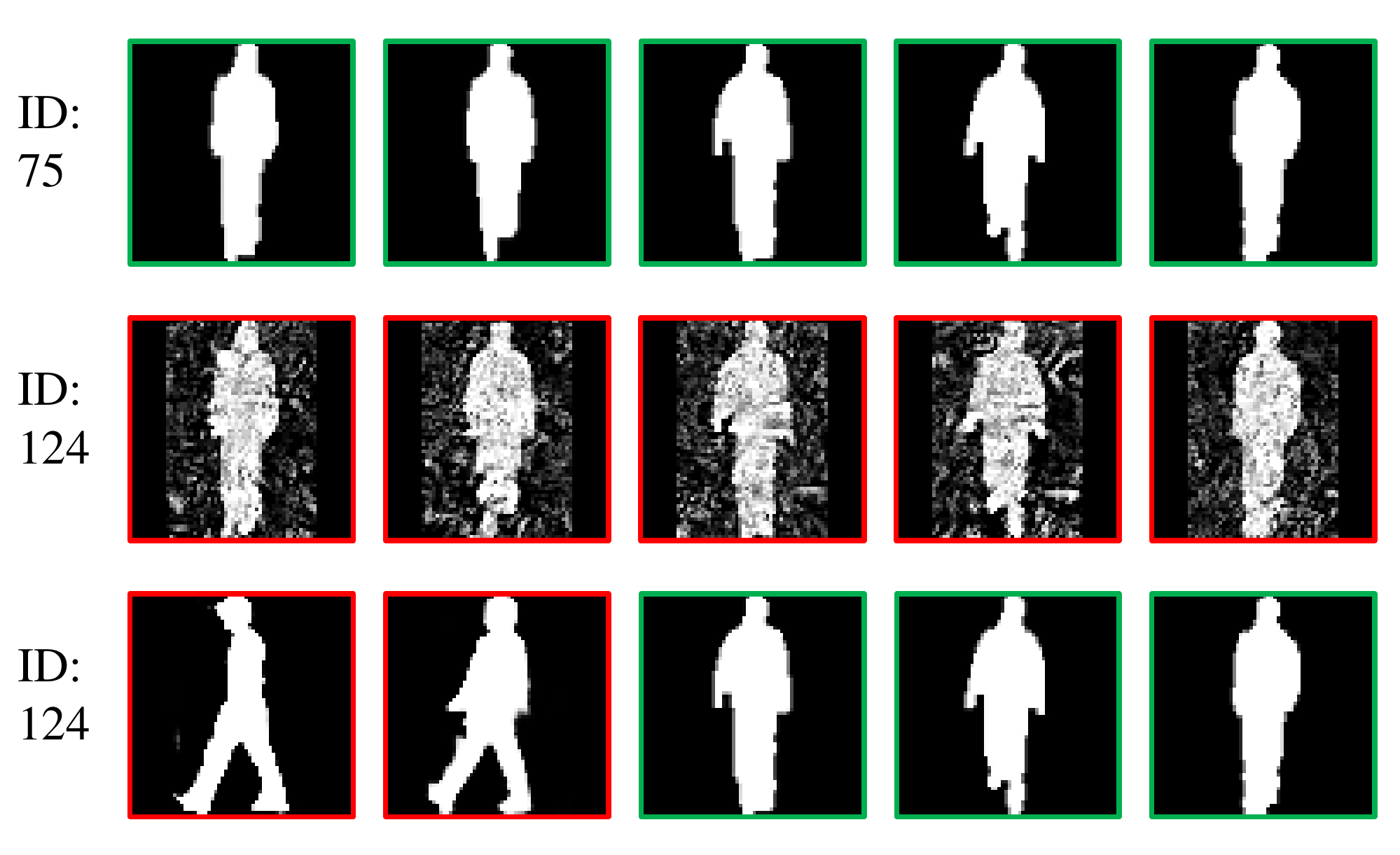}
		\caption{\textbf{Top row}: the original examples. \textbf{Middle row}: the perturbation-based adversarial examples. \textbf{Bottom row}: the temporal sparse adversarial examples. The red bounding box represents the modified example, while the green bounding box means the original example. The middle row directly transfers adversarial attack methods in image classification to gait recognition, causing all frames perturbed and imperceptibility decreased. The bottom row has only the first two frames modified. Besides, the modified frames maintain a gait appearance, so it is not easy to distinguish whether they are adversarial examples or not from human vision.}
		\label{fig:1}
	\end{figure}

	%comment: consider clarifying "performance"
	In this paper, we investigate the security of gait recognition models subjected to adversarial attacks~\cite{Szegedy2013IntriguingPO,Goodfellow2014ExplainingAH}. Different from typical spoofing attacks, adversarial attacks aim to imperceptibly (i.e., without incurring visual cues) disable the gait recognition model (nontargeted attack) or gain illegitimate access by misleading the system to recognise as a target ID (targeted attack). 
	%, not to mention adversarial attacks which can fool most machine learning models without causing visual awareness. 
	Recently, adversarial attacks have been investigated including attacks on image classification~\cite{Szegedy2013IntriguingPO,Goodfellow2014ExplainingAH}, object detection~\cite{Xie2017AdversarialEF}, face recognition~\cite{Sharif2016AccessorizeTA}, etc. However, for gait recognition, to the best of our knowledge, a meaningful attempt has not been reported, yet. A very likely reason is that the popular adversarial attack methods on image classification, like Fast Gradient Sign Method (FGSM)~\cite{Goodfellow2014ExplainingAH} and Projected Gradient Descent (PGD)~\cite{Madry2017TowardsDL}, are not suitable to directly applied to gait recognition. Firstly, for sequence-based methods that take a sequence of silhouettes segmented from the original video as input, perturbations added on the source video do not work. This is due to the signal processing these approaches require. Secondly, even if attackers have access to modify the probes, adding a norm-constrained perturbation to the original gait silhouette destroys the imperceptibility. This is illustrated in the second row of Fig.~\ref{fig:1}.
	
	\iffalse
	\begin{figure}
		\centering
		\includegraphics[width=1.0\linewidth]{fig0}
		\caption{\textbf{Top row}: the original sequence. \textbf{Middle row}: spoofing attack. \textbf{Bottom row}: adversarial attack.}
		\label{fig:0}
	\end{figure}
	\fi
	
	More specifically, this work focuses on the sequence-based methods~\cite{Wolf2016MultiviewGR,Liao2017PoseBasedTN,Chao2018GaitSetRG}. Compared to template-based methods~\cite{Wu2016ACS,Takemura2019OnIA,He2019MultiTaskGF}, in which temporal information is difficult to preserve, sequence-based methods are better at extracting dynamic clues from silhouette frames with deep neural networks (DNNs). As a result, these methods have a higher gait recognition accuracy. However, the DNN-extracted temporal features may be vulnerable to adversarial attacks. To verify this hypothesis, we propose a novel temporal sparse adversarial attack method for the gait recognition system. 
	%One natural idea is to directly apply the popular adversarial attack methods on image classification, like Fast Gradient Sign Method (FGSM)~\cite{Goodfellow2014ExplainingAH} and Projected Gradient Descent (PGD)~\cite{Madry2017TowardsDL}. However, these perturbation-based methods are not entirely suitable for the gait recognition task for two reasons.
	%comment: consider introducing the acronym DNN

	%comment: consider explicitly introducing your hypothesis
	
	%The main difference between adversarial attacks against sequence-based gait recognition and other tasks, such as image classification or face recognition, is the imperceptibility of adversarial examples. 

	%For the above reasons, we propose a novel temporal sparse adversarial attack on gait recognition. 
	We have two primary intuitions that are illustrated in Fig.~\ref{fig:1}.
	Firstly, the input of gait recognition models is a sequence of silhouette frames, rather than a single image for the image classification models. Therefore, to better achieve its imperceptibility, only a few frames may be modified in our attack. This ensures sparsity on the temporal domain. Secondly, motivated by unrestricted adversarial examples~\cite{Brown2018UnrestrictedAE}, crafting an unrestricted adversarial gait silhouette via deformation better achieves imperceptibility than adding norm-bounded perturbations. Moreover, adversarial silhouettes generated by the proposed method can easily be extended to valid video frames. This enables a practical threat to gait recognition systems. %Our method disables the detection methods of adversarial perturbations%We show a simple illustration in Fig.~\ref{fig:1}. %Someone may try to add perturbations to the source video. But for silhouette-based methods that take a sequence of silhouettes segmented from the original video as input, it still does not work. For these two reasons, we propose our temporal sparse adversarial attack on gait recognition. 
	
	%natural images of different categories are easy to distinguish by humans eyes, whereas the gait silhouette of different subjects is not that case. Thus
	
	%We first train a generator to generate gait silhouette utilizing the popular WGAN-GP~\cite{Gulrajani2017ImprovedTO} and then optimize stochastic latent variables which are inputted into the generator utilizing the adversarial attack algorithm MIFGSM~\cite{Dong2017BoostingAA}. Finally, we input the optimal latent variables into the generator to obtain adversarial examples. %As shown in [], our adversarial examples has better imperceptibility and meanwhile achieve high attack success rate on the state-of-the-art gait recognition model.
	
	The main contributions of this paper are as follows:
	\begin{itemize}
		%\vspace{-1ex}
		\item[(1)] %To the best of our knowledge, we are the first attempt to address adversarial attacks on sequence-based gait recognition. %Experiments show the vulnerability of gait recognition systems.
		%\vspace{-1ex}
		%\item To evaluate the attack intensity properly, we apply the $l_{2,1}$ norm to define a new evaluation criterion.
		%\item 
		%To evaluate the attack intensity properly, we define a new evaluation criterion by combining the $l_{1}$ norm in the time domain and semantic similarity in the spatial domain. 
		%\vspace{-1ex}
		%\item We propose a novel temporal sparse adversarial attack method on gait recognition. Our attack achieves a high attack success rate and good imperceptibility simultaneously.
		We propose a novel temporal sparse adversarial attack specifically designed to target gait recognition methods. The proposed method simultaneously achieves a high attack success rate and satisfactory imperceptibility.
		\item[(2)] We conduct extensive experiments to study the vulnerability of existing sequence-based gait recognition systems. %We also investigate the difference between sequence-based and template-based gait recognition methods under our proposed attack method. 
		The results indicate that sequence-based deep learning methods have little adversarial robustness despite their high accuracy.
		
		%commnet: clarify performance (assumed accuracy in the editing)
		
	\end{itemize}
	
	\section{Related work}
	\label{sec:relate}
	\subsection{Gait recognition}
	\label{subsec:gait}
	Gait recognition can generally be grouped into two categories, template-based~\cite{Wu2016ACS,Takemura2019OnIA,He2019MultiTaskGF} and sequence-based~\cite{Wolf2016MultiviewGR,Liao2017PoseBasedTN}. The former category is composed of two main steps: template generation and matching. In the first step, human silhouettes are compressed into one template. For example, GEINet~\cite{DBLP:conf/icb/ShiragaMMEY16} and GaitGAN~\cite{DBLP:conf/cvpr/YuCRP17} use the gait energy image (GEI)~\cite{Han2006IndividualRU} as the template. In the second step, the similarity between pairs of templates is evaluated, e.g., by the Euclidean distance. The latter category directly captures dynamic clues from the sequence of silhouette frames. This category includes 3D CNN-based approaches~\cite{Wolf2016MultiviewGR}, LSTM-based approaches~\cite{Liao2017PoseBasedTN}, and GaitSet~\cite{Chao2018GaitSetRG}. Currently, GaitSet achieves the state-of-the-art gait recognition results on the CASIA-B~\cite{Yu2006AFF} dataset. 
	
	% comment: consider clarifying "better"
	
	\subsection{Adversarial attack}
	\label{subsec:adv}
	Let $x \in \mathbb{R}^m$ denote an input to a classifier $f : \mathbb{R}^m \rightarrow \{1, 2, ... , k\}$, and assume the attacker has full knowledge of $f$. %In my opinion, related work here is clear.\Wei{where $m$ is ..., and $k$ is ...}. 
	The goal of a nontargeted adversarial attack is to find the corresponding adversarial example  $x^{\ast}\in \mathbb{R}^m$ satisfying $f(x^{\ast}) \neq f(x)$ with the constraint $\|x^{\ast}-x\|\leq \epsilon$, where $\epsilon$ is a small constant. For targeted attack, the $x^{\ast}\in \mathbb{R}^m$ aims to satisfy $f(x^{\ast})=t$, where $t$ is the target label. In this additive perturbation approach, different norms $\|\cdot\|$ have been used, such as $l_1$~\cite{Kurakin2016AdversarialML}, $l_2$~\cite{MoosaviDezfooli2015DeepFoolAS} or $l_0$~\cite{Papernot2015TheLO}. A series of methods have been proposed such as FGSM~\cite{Goodfellow2014ExplainingAH},  PGD~\cite{Madry2017TowardsDL}, and MIFGSM~\cite{Dong2017BoostingAA}. The adversarial perturbations are typically restricted to a small norm. 
	
	In contrast, unrestricted adversarial examples~\cite{Brown2018UnrestrictedAE,Song2018ConstructingUA,Poursaeed2019FinegrainedSO} are constructed entirely from scratch instead of perturbing existing data points by a small amount. %Some methods~\cite{Song2018ConstructingUA,Poursaeed2019FinegrainedSO} focus on generating adversarial examples of this special type. 
	Poursaeed et al.~\cite{Poursaeed2019FinegrainedSO} manipulate stylistic and stochastic latent variables that are fed into the StyleGAN~\cite{Karras2018ASG} to generate an unrestricted adversarial image to mislead a classification model. 
	Similarly, we adopt a generative model to generate an adversarial high-quality gait silhouette. Here, we extend the approach to include the temporal domain. Instead of perturbing each frame, we sparsely generate adversarial frames to alert or insert into the original gait sequence. %A temporal mask is employed to control the sparsity of the adversarial frames. 
	
	In addition, additional approaches are available in the literature on adversarial attacks via action recognition~\cite{Wei2018SparseAP,Chen2019AppendingAF}. Wei et al.~\cite{Wei2018SparseAP} utilize $l_1$ norm across frames to ensure the sparsity of adversarial perturbations on videos. A similar mask-based method is applied in our attack to control the sparsity. 
	Chen et al.~\cite{Chen2019AppendingAF} propose a new adversarial attack that appends a few dummy frames to a video clip and then adds adversarial perturbations only on these new frames. In our attack, we also explore the strategy of inserting frames into gait sequences. %Nonetheless, b
	Both methods~\cite{Wei2018SparseAP,Chen2019AppendingAF} achieve a superior success rate on attacking temporal sequences. Nonetheless, their methods focus on the norm-bounded perturbations and cannot be directly transferred to the gait recognition task for the reason presented in Sec.~\ref{sec:intro}.%Both the methods focus on the norm-bounded perturbations and need to tune the hyperparameters in the objective. In our task, we relax the constraint of perturbations and simplify our objective to omit the tuning.  %\Wei{Clarify the difference with these two methods.} 
	
	%comment: consider clarifying "performance"
	
	%We demonstrate with our method, inserting the generated adversarial gait images without norm-bounded perturbations into an original sequence can also perform a successful attack.
	
	%All these methods are most widely used in deep image classification models,

	\subsection{Generative adversarial network}
	Image generation is highly related to our task. A core challenge is generating silhouettes or video frames that are visually realistic. Generative adversarial network (GAN)~\cite{Goodfellow2014GenerativeAN} achieves impressive results in image synthesis and thus is applied in our method. Wasserstein GAN (WGAN)~\cite{DBLP:journals/corr/ArjovskyCB17} is an important extension of GAN which improves image quality and stabilizes training. WGAN-GP~\cite{DBLP:conf/nips/GulrajaniAADC17} uses a gradient
	penalty to further improve the loss function. We use the WGAN-GP in the silhouette generation. 
	
	Another related study is the pixel-to-pixel generation approach ~\cite{DBLP:conf/cvpr/IsolaZZE17}, which is based on a type of conditional GANs for which both input and output are images. SPADE~\cite{DBLP:conf/cvpr/Park0WZ19} is a method based on conditional normalization, and it can convert the segmentation map into a photo-realistic image. This is suitable for our silhouette to video frame generation. In this paper, although the adversarial frames are hard to detect due to the temporal sparsity, we still generate visually convincing frames for better imperceptibility in the spatial domain. Successfully fooling recognition systems in such a case demonstrates the effectiveness of our method.
	
	\begin{figure*}
		\centering
		\includegraphics[width=1.0\linewidth]{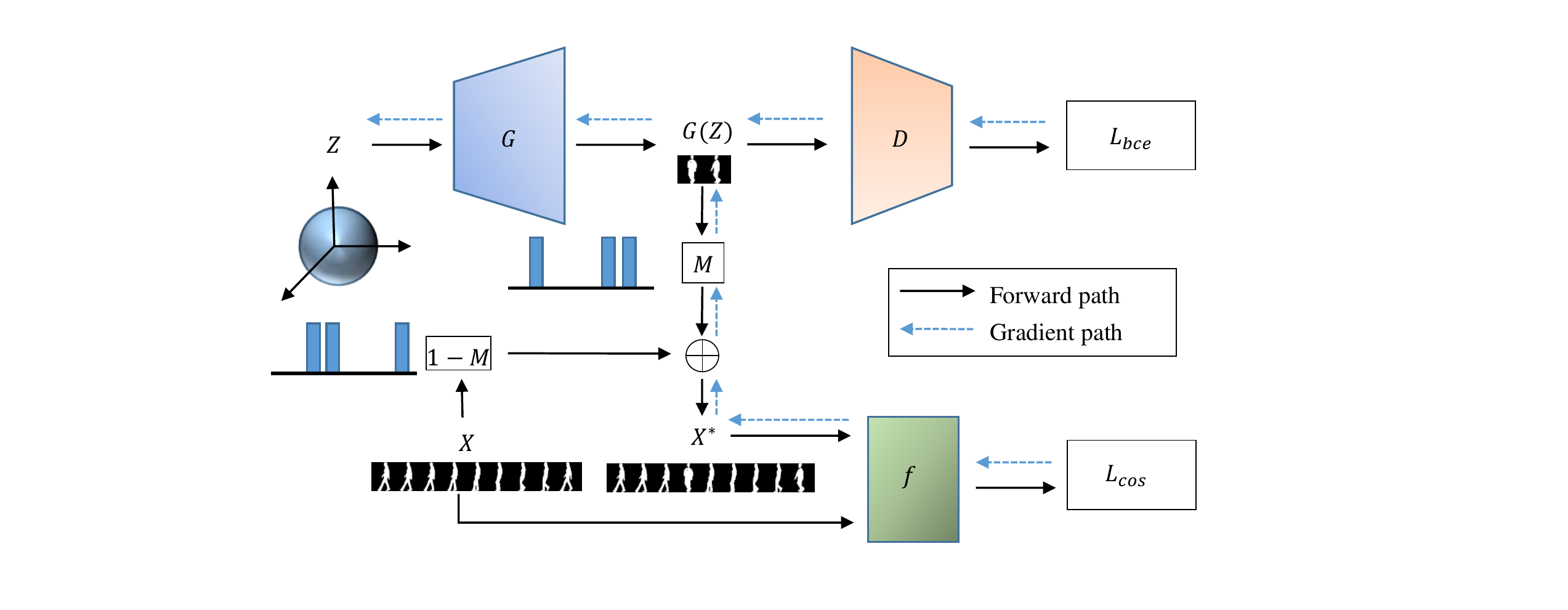}
		\caption{The pipeline of our attack approach. During attacking process, the latent vector $\bm{Z}$ is the only parameter to be optimized. We start from a randomly sampled vector and iteratively optimize it through gradient backpropagation, shown as the blue dotted arrows. Finally, we feed the optimized latent vector into the generator $G$, and then put the obtained adversarial silhouettes into a source sequence to fool the target gait model $f$. The discriminator $D$ is used to supervise high-quality gait silhouette image generation for imperceptibility on frame level. The mask $M$ is used to achieve temporal sparsity for imperceptibility on the sequence level.}
		\label{fig:model}
	\end{figure*}
	
	\section{Methodology}
	\label{sec:method}
	
	\subsection{Problem formulation}
	%The adversarial gait frame is expected to be as similar as normal gait silhouette. In ~\cite{Wei2018SparseAP}, Wei et al. study the adversarial perturbations for videos and they use a temporal mask on the video to better control the sparsity. The difference is our task is focus on an unordered set rather than a sequence with temporal imformation such that the mask can be added on any frame.
	
	%We first train a generator on a gait dataset and denote the trained generator as G. 
	
	Let $\bm{X} \in \mathbb{R}^{N \times W\times H\times C}$ denote a clean silhouette sequence, and $\bm{X^{\ast}}\in \mathbb{R}^{N \times W\times H\times C}$ denote its adversarial sequence, where $N$ is the number of frames, and $W, H, C$ are the width, height, and channel for a specific frame, respectively. 
	
	The nontargeted adversarial sequence $\bm{X^{\ast}}$ is the solution of the following objective function:
	\begin{equation}
	\label{eq:objective}
	%\argmin_{Z} - \mathcal{L}_{cos}(f({X}),f(X^{\ast}))
	\argmin_{\bm{X^{\ast}}} \lambda \mathcal{C}(\bm{X},\bm{X^{\ast}}) - \mathcal{L}_{cos}(f({\bm{X}}),f(\bm{X^{\ast}})),
	\end{equation}
	where $\lambda$ is a weight that balances the two terms in the objective function and $f$ is a gait recognition model that outputs the computed features of silhouette sequences. In addition, $\mathcal{L}_{cos}$ is the loss function to measure the cosine similarity between the ground truth sequence and the adversarial sequence; $\mathcal{C}(\bm{X},\bm{X^{\ast}})$ is a distortion measurement to evaluate the difference between the original sequence and its adversarial sequence. For a perturbation-based attack, it is often defined as the $l_p$ norm $\|\bm{X^{\ast}}-\bm{X}\|_p$. For our attack, we define a new measurement as 
	\begin{equation}
	\label{eq:creterion}
	%\argmin_{Z} - \mathcal{L}_{cos}(f({X}),f(X^{\ast}))
	\mathcal{C}(\bm{X},\bm{X^{\ast}}) = \sum_{n\in\varPhi}  (o(X_n)-o(X_n^{\ast}))^2,
	\end{equation}
	where $o$ is the oracle to decide whether the image is a reasonable gait silhouette and similar to its counterpart. As unrestricted adversarial examples~\cite{Brown2018UnrestrictedAE}, the adversarial frames in our attack are expected to maintain a gait appearance even though they may have a large perturbation at the pixel-level. $\varPhi$ is a subset within the set of frame indices. %For any $n\in \varPhi$, we have $X_n\neq X_n^{\ast}$, which means the $n$-th frame in adversarial sequence is different from the $n$-th frame in original sequence. 
	
	For clarity, we describe our method based on nontargeted setting. Our method can be easily generalized to targeted setting. The objective for finding an adversarial sequence misclassified as a target ID $t$ is to minimize the loss function 
	\begin{equation}
	\label{eq:taret_objective}
	%\argmin_{Z} - \mathcal{L}_{cos}(f({X}),f(X^{\ast}))
	\argmin_{\bm{X^{\ast}}} \lambda \mathcal{C}(\bm{X},\bm{X^{\ast}}) + \mathcal{L}_{cos}(f({\bm{X_t}}),f(\bm{X^{\ast}})),
	\end{equation}
	where $\bm{X_t}$ is a sequence of ID $t$.
	
	\begin{figure*}
		\centering
		\includegraphics[width=0.8\linewidth]{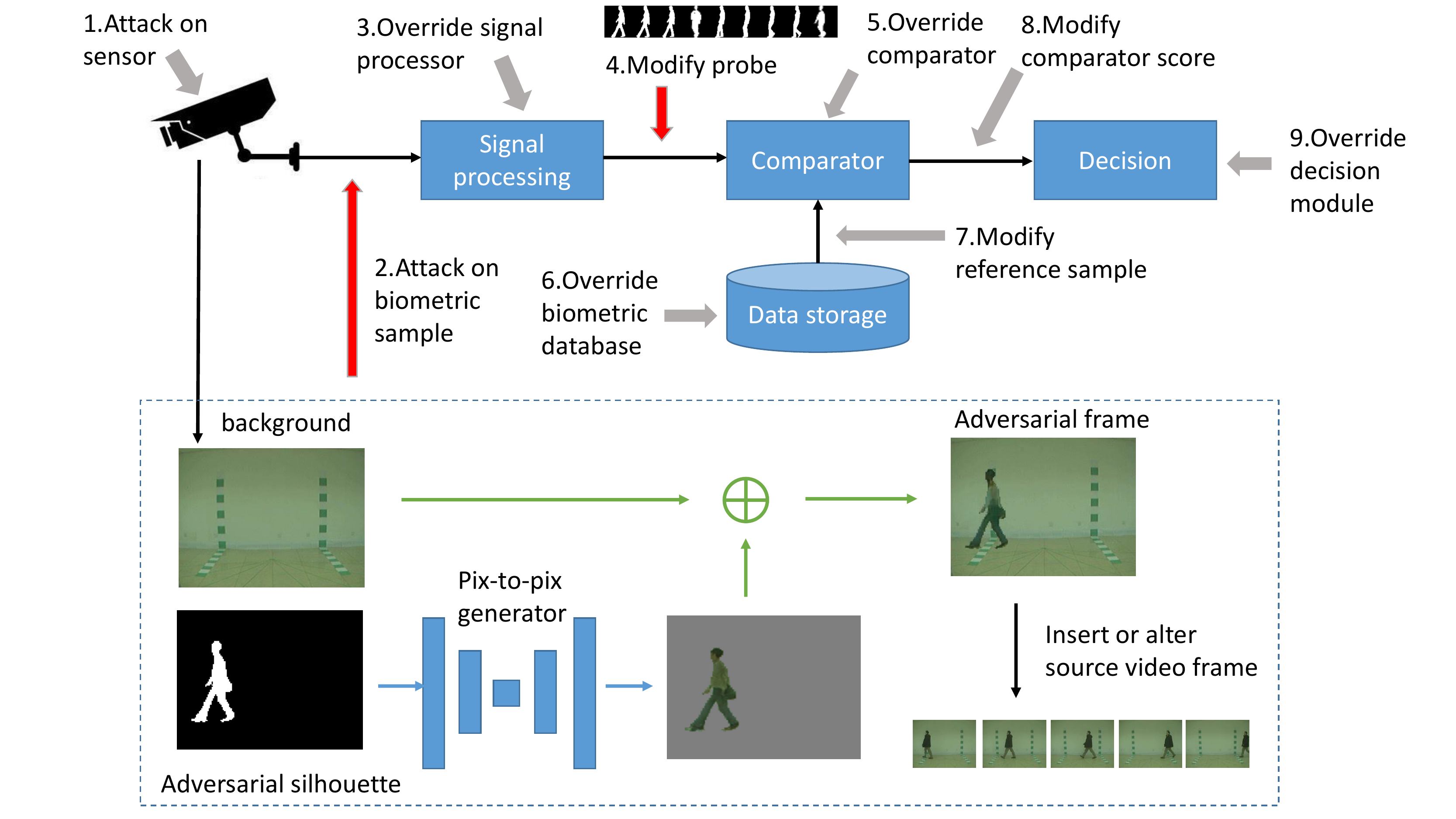}
		%\captionsetup{font={footnotesize}}
		
		\caption{Vulnerability of a biometric recognition system. This figure is inspired by Jia et al.~\cite{Jia2019AttackingGR}. Our method can be applied in both the 2nd and 4th step. To attack on biometric samples, we train a generator to translate an adversarial silhouette image into a valid video frame.}
		
		\label{fig:system}
	\end{figure*}
	
	\subsection{Temporal sparse attack}
	
	%Due to the new criterion in Eq.(\ref{eq:creterion}), directly optimizing the objective in Eq.(\ref{eq:objective}) is not a good choice. Thus we propose our new attack method, the pipeline of which is shown in Fig.~\ref{fig:model}.
	
	In our new measurement (\ref{eq:creterion}), the supervision of the oracle is of vital importance to improve the imperceptibility in the spatial domain. However, the objective function Eq.(\ref{eq:objective}) is difficult to optimize. Thus, we use a well-trained generator to craft adversarial frames as in Poursaeed et al.~\cite{Poursaeed2019FinegrainedSO}. The pipeline of our attack is shown in Fig.~\ref{fig:model}.

	%\subsection{Alter-frame Attack}
	
	To control the temporal sparsity, similar to ~\cite{Wei2018SparseAP}, we denote the temporal mask as $\bm{M}\in\{\bm{0},\bm{1}\}^{N \times W\times H\times C}$. We let $\varOmega =\{1,2,...,N\}$ be the set of frame indices, $\varPhi$ be a subset within $\varOmega$ having $K$ elements randomly sampled from $\varOmega$, and $\varPsi=\varOmega-\varPhi$. The selection of $\varPhi$ introduces randomization to make the crafted adversarial sequence more difficult to detect. If $n\in \varPhi$, we set $M_n = 0$, and if $n\in \Psi$, $M_n = 1$, where $M_n \in \{\bm{0},\bm{1}\}^{W\times H\times C}$ is the n-th frame in $M$. The sparsity is computed as $S = K/N$. 
	
	Denote the latent variables input into the generator as $\bm{Z}\in \mathbb{R}^{N \times V}$, where $V$ is the dimension of each latent variable. $G$ is a pre-trained generator on a gait silhouette dataset. Let $\mathcal{M}$ be the natural gait silhouette manifold in $\mathbb{R}^{W\times H}$. In most generative models, a simple random sample $\bm{Z}$ drawn from the standard Gaussian distribution does not guarantee that $G(\bm{Z})\in \mathcal{M}$. To ensure the high quality of generated silhouettes, we must be in a region of the latent space with high probability. Inspired by Menon et al.~\cite{PULSE}, we replace the Gaussian prior on $\mathbb{R}^V$ with a uniform prior on $\sqrt{V}\mathbb{S}^{V-1}$, where $\mathbb{S}^{V-1}\subset \mathbb{R}^V$ is the unit sphere in the $V$ dimensional Euclidean space. The adversarial sequence is obtained by
	\begin{equation}
	\bm{X^{\ast}}=\bm{M}\cdot G(\bm{Z})+(\bm{1}-\bm{M})\cdot \bm{X}.
	\end{equation}
	%where $G$ is a trained generator on a gait silhouette dataset.
	We name this method as a \textbf{frame-alteration attack}. Instead of modifying frames, a \textbf{frame-insertion attack} inserts frames into the original sequence to obtain the adversarial sequence $\bm{X^{\ast}}\in \mathbb{R}^{(N+\Delta N) \times W\times H\times C}$. 
	
	%comment: consider clarifying "few"
	
	For the well trained generator $G$, all of the generated images $G(\bm{Z})$ must be classified by the oracle as gait silhouettes, i.e., $\mathcal{C}(\bm{X},\bm{X^{\ast}})=0$. To ensure this, we use a trained discriminator $D$ to supervise the generated silhouettes. The discriminator outputs the value one when the inputs are from the natural manifold; otherwise, the value zero is output. We make sure $G(\bm{Z})$ keeps a high probability of sampling from the natural manifold by utilizing the binary cross entropy loss $\mathcal{L}_{bce}$.
	Thus, our objective function in Eq.(\ref{eq:objective}) is equal to optimizing $\bm{Z}$ to maximize the following loss:
	\begin{equation}
	\label{eq:newobjective}
	\mathcal{L} = \mathcal{L}_{cos}(f({\bm{X}}),f(\bm{X^{\ast}})) + \lambda\mathcal{L}_{bce}(D(G(\bm{Z})),\bm{1}),
	%\argmax_{\bm{Z}} %\mathcal{L}_{cos}(f({\bm{X}}),f(\bm{X^{\ast}}))
	%\argmin_{X^{\ast}} \lambda C(X,X^{\ast}) - \mathcal{L}_{cos}(f({X}),f(X^{\ast}))
	\end{equation}
	where $\lambda$ is a hyper-parameter to establish a trade-off between two terms. By performing a gradient ascent in the latent variable space of the generator, the corresponding $\bm{Z}$ that maximizes the final loss in Eq.(\ref{eq:newobjective}) can be found. Without loss of generality, we adopt the MIFGSM~\cite{Dong2017BoostingAA} to attack $f$ as follows:
	\begin{equation}
	\label{eq:mifgsm1}
	\bm{g}_{(t+1)} = \mu\cdot \bm{g}_{(t)}+\frac{\bigtriangledown_{\bm{Z}_{(t)}}\mathcal{L}}{\|\bigtriangledown_{\bm{Z}_{(t)}}\mathcal{L}\|_1},
	%\bm{g}_{(t+1)} = \mu\cdot \bm{g}_{(t)}+\frac{\bigtriangledown_{\bm{Z}_{(t)}}\mathcal{L}_{cos}(f({\bm{X}}),f(\bm{X^{\ast}}_{(t)}))}{\|\bigtriangledown_{\bm{Z}_{(t)}}\mathcal{L}_{cos}(f({\bm{X}}),f(\bm{X^{\ast}}_{(t)}))\|_1}
	\end{equation}
	\begin{equation}
	\label{eq:mifgsm2}
	\bm{Z}_{(t+1)} = \bm{Z}_{(t)}+\epsilon\cdot sign(\bm{g}_{(t+1)}),
	\end{equation}
	where $\mu$ is the decay factor, $\epsilon$ is the step size, and $t$ represents the $t$-th iteration. Algorithm~\ref{alg:attack} shows the proposed temporal sparse adversarial attack.
	
	%We conclude our method in Algorithm~\ref{alg:attack} and it can be easily extended to another type of attack --- insert-frame attack. Instead of modifying a few frames, insert-frame attack insert a few frames into the origin sequence to obtain the adversarial sequence $X^{\ast}\in \mathbb{R}^{(N+\Delta N) \times W\times H\times C}$. Then the MIFGSM algorithm is used to solve the same problem in Eq.(\ref{eq:objective}). 
	
	\begin{algorithm}
		\small
		\caption{Temporal Sparse Adversarial Attack}
		\label{alg:attack}
		\begin{algorithmic}[1]
			\Require A gait recognition model $f$; a generator $G$; a discriminator $D$; a silhouette sequence $\bm{X}$; iterations $T$ and decay factor $\mu$; sparsity $S$; step size $\epsilon$; latent space dimension $V$; a hyper-parameter $\lambda$.
			\Ensure
			An adversarial silhouette sequence $\bm{X}^*$.
			\State $\bm{g}_0 = 0$; $\bm{X}_0^* = \bm{X}$; %$\bm{Z}_0\sim Gaussian(0,1)$.
			$\bm{Z}_0\sim \sqrt{V}\mathbb{S}^{V-1}$, where $\mathbb{S}^{V-1}$ is the unit sphere space.
			\State Compute the mask $\bm{M}$ according to the sparsity $S$, details are in the text;
			\For {$t = 0$ to $T-1$}
			\State Input $\bm{Z}_t$ into the generator $G$ and obtain the images $G(\bm{Z}_t)$;
			\State Compute the adversarial sequence as $\bm{X}^{\ast}_t=\bm{M}\cdot G(\bm{Z}_t)+(1-\bm{M})\cdot \bm{X}$;
			\State Compute the %cosine similarity loss of two sequences features as $\mathcal{L}_{cos}(f({\bm{X}}),f(\bm{X}^{\ast}_t))$;
			loss $\mathcal{L}$ in Eq.(\ref{eq:newobjective});
			\State Update $\bm{g}_{(t+1)}$ by accumulating the velocity vector in the gradient direction as 
			%\vspace{-2ex}
			Eq.(\ref{eq:mifgsm1});
			%\vspace{-3ex}
			\State Update $\bm{Z}_{(t+1)}$ by applying the clipped gradient as
			%\vspace{-1.5ex}
			Eq.(\ref{eq:mifgsm2});
			%\vspace{-4ex}
			\EndFor \\
			\Return $\bm{X}^* = \bm{M}\cdot G(\bm{Z}_T)+(1-\bm{M})\cdot \bm{X}$.
		\end{algorithmic}
	\end{algorithm}

	\subsection{Video generation}
	The above generation process only considers silhouette images. To show our method can successfully threaten other practical applications, we extend it to the generation of a valid video. This can be regarded as a pixel-to-pixel image generation task. Specifically, we apply the popular SPADE~\cite{DBLP:conf/cvpr/Park0WZ19} in our experiments. The whole pipeline is shown in Fig.\ref{fig:system}.

	%comment: consider providing a very brief description of SPADE (e.g., one or two sentences)
	
	Denote the source silhouette as $L_s\in \mathbb{R}^{W\times H}$ and the source video frame as $I_s\in \mathbb{R}^{W\times H}$. We train a pix-to-pix generator $G_{p}$ with paired data $(I_s\times L_s, L_s)$. In the attack process, we feed the adversarial silhouette $L_a$ into the generator. The generated image $G_{p}(L_a)$ is supposed to contain a subject; there is no background scene. We paste the generated image on the background image $I_b$ with the formulation $I_a = I_b \times L_s + G_{p}(L_a)$. Then, we insert the obtained adversarial frame into the source video or substitute a frame with it to generate the fake video. Though this method does not ensure spatial-temporal continuity between the adjacent real frames and adversarial frames, the modified frames are imperceptible due to the temporal sparsity. On the other hand, the modified frame keeps the same background, which is enough to deceive segmentation algorithms such as background difference. Moreover, some sequence-based models, like GaitSet, are flexible and capable of containing non-consecutive silhouettes in input sets. Thus, in these scenarios our temporal sparse adversarial video is not easy to detect; our video can cause a real threat to practical applications.

	\section{Experiments}
	\label{sec:exp}
	%To validate the proposed method, we conduct extensive experiments on the CASIA-B[] dataset. 
	In this section, we conduct experiments to explore the vulnerability of gait recognition models under our temporal sparse attack. First, we specify the experimental settings in Sec.~\ref{sec:setup}. Then, we test the adversarial robustness of the state-of-the-art gait recognition model via the proposed method. The white-box attack is presented in Sec.~\ref{sec:results} and a cross-dataset validation is presented in Sec.~\ref{sec:cross}. We also perform a black-box attack to investigate whether adversarial examples of sequence-based models can transfer to template-based models in Sec.~\ref{sec:black}. Moreover, we provide a comparison of the proposed method with existing perturbation-based methods to demonstrate the superiority of our method in Sec.~\ref{sec:compare}. Finally, we make a further analysis of the proposed method in Sec.~\ref{sec:analysis}.
	
	\subsection{Setup}
	\label{sec:setup}
	
	{\bf Datasets.}
	%The dataset we conduct experiments on is CASIA-B~\cite{Yu2006AFF}.
	We conduct experiments on two datasets, CASIA-A~\cite{Yu2006AFF} and CASIA-B~\cite{Yu2006AFF}. CASIA-A consists of 20 subjects, and each subject has 12 image sequences, 4 sequences for each of the three directions, i.e. parallel, 45 degrees and 90 degrees to the image plane. Each sequence is labeled with `mm$\_$n', where `mm' represents direction and `n' is sequence number. For example, 4 parallel sequences are labeled with $00\_1$, $00\_2$, $00\_3$, $00\_4$, respectively. %$45\_1$, $45\_2$, $45\_3$, $45\_4$, $90\_1$, $90\_2$, $90\_3$, and $90\_4$. 
	CASIA-B is a widely used gait dataset that contains 124 subjects (labeled 001-124) with 11 different viewing angles and 10 sequences per subject for each view. The 10 sequences contain three walking conditions: six sequences are in the normal walking state (NM 1-6), two sequences contain walking subject wearing coats (CL 1-2), and two sequences contain subject carrying bags (BG 1-2). We mainly use CASIA-B for evaluation and CASIA-A for cross-dataset validation. 
	
	Since our target gait models are trained on the first 74 subjects (labeled 1-74) and tested on the remaining 50 subjects of CASIA-B, we follow this setting to attack the last 50 subjects (labeled 75-124). For each subject, the first four sequences of the NM condition (NM 1-4) are kept in the gallery to test the recognition accuracy. All of the frames in a specific view and walking condition are used as a sequence for the test. For the cross-dataset validation, we use the whole CASIA-A. The first three sequences of each angel are in the probe and the fourth sequence is in the gallery. The setting is summarized in Table~\ref{tab:setting}.

	\begin{table}
		\caption{The dataset setting.}
		\label{tab:setting}
		\begin{center}
			\setlength{\tabcolsep}{3mm}
			%\resizebox{75mm}{40mm}{
			%\resizebox{\columnwidth}{40mm}{
			\renewcommand{\arraystretch}{1.1}
			\begin{tabular}{|c|c|c|c|}
				\hline
				\multirow{12}*{CASIA-B} & \multicolumn{2}{c|}{\multirow{4}{*}{training set}} & 
				ID: 001-074\\
				& \multicolumn{2}{c|}{}&nm01-nm06,\\
				& \multicolumn{2}{c|}{}&bg01-nm02,\\
				& \multicolumn{2}{c|}{}&cl01-cl02,\\
				\cline{2-4}
				& \multicolumn{2}{c|}{\multirow{2}{*}{gallery set}} & ID: 075-124 \\
				& \multicolumn{2}{c|}{}& nm01-nm04\\
				\cline{2-4}
				& \multirow{6}{*}{probe set} & \multirow{2}{*}{probeNM}& ID: 075-124\\
				&  & \multirow{2}{*}{}& nm05,nm06\\
				\cline{3-4}
				& & \multirow{2}{*}{probeBG}& ID: 075-124\\
				&  & \multirow{2}{*}{}& bg01,bg02\\
				\cline{3-4}
				& & \multirow{2}{*}{probeCL}& ID: 075-124\\
				&  & \multirow{2}{*}{}& cl01,cl02\\
				\hline
				\multirow{8}{*}{CASIA-A} & \multicolumn{2}{c|}{\multirow{2}{*}{gallery set}} & ID:all\\
				& \multicolumn{2}{c|}{\multirow{2}{*}{}} & 00\_4, 45\_4, 90\_4\\
				\cline{2-4}
				& \multirow{6}{*}{probe set} & \multirow{2}{*}{probe $0^{\circ}$}& ID:all\\
				&  &  & 00\_1, 00\_2, 00\_3\\
				\cline{3-4}
				&  & \multirow{2}{*}{probe $45^{\circ}$}& ID:all\\
				&  & &  45\_1, 45\_2, 45\_3\\
				\cline{3-4}
				&  & \multirow{2}{*}{probe $90^{\circ}$}& ID:all\\
				&  & &  90\_1, 90\_2, 90\_3\\
				\hline
			\end{tabular}
			%}
		\end{center}
		
	\end{table}
	
	\iffalse
	{\bf Models.} The attacked model is GaitSet, the state-of-the-art gait recognition model. GaitSet regards the gait as a set of gait silhouettes and utilizes a deep neural network to directly extract temporal information during training. Moreover, GaitSet is flexible since the input set can contain any number of non-consecutive silhouettes.
	
	For the black-box experiments, we use the template-based model GaitGAN~\cite{DBLP:conf/cvpr/YuCRP17}. Different from GaitSet, which takes a gait sequence as a set and extracts its feature with a CNN, GaitGAN uses a GEI template as the gait feature. Moreover, GaitGAN takes a GAN model as a regressor to simultaneously address variations in viewpoint, clothing, and carrying conditions in gait recognition.
	
	The order of input frames does not affect the recognition accuracy in GaitSet and GaitGAN. To study what effect the positions of adversarial frames have on the attacking performance, we use SelfGait~\cite{liu2021selfgait}, a self-supervised framework with spatiotemporal components to learn from the massive unlabeled gait images. SelfGait preserves and learns temporal representation from the order and relationship of frames in gait sequences. Disrupting the order of input frames will decrease its accuracy.
	\fi
	
	{\bf Metrics.} We evaluate the vulnerability of gait recognition models by assessing their accuracy. The accuracy is averaged on all gallery views, and the identical views are excluded. For example, when testing with CASIA-B, the accuracy of the probe view
	$90^{\circ}$ is averaged on 10 gallery views, excluding gallery view $90^{\circ}$. 
	
	{\bf Implementation details.} If not specifically mentioned, we conduct nontargeted attack. For targeted attack, we randomly choose the target ID and select one sequence of targeted ID as $\bm{X_t}$ in Eq.(\ref{eq:taret_objective}). 
	
	For the mask $\bm{M}$, we let the set $\varPhi$ be $\varPhi = \{1,2,...,K\}$, which means we simply alter the first $K$ frames with adversarial ones. $K$ is computed according to the needed sparsity. 
	
	WGAN-GP~\cite{DBLP:conf/nips/GulrajaniAADC17} is an important extension of GAN which improves image quality and stabilizes training. We train the WGAN-GP on CASIA-B for 16000 iterations; the trained generator $G$ and the discriminator $D$ are used in our attack. %WGAN-GP is an important extension of GAN which improves image quality and stabilizes training. 
	We set the dimension of inputted latent variables as $V=128$.
	
	Our attack is based on MIFGSM~\cite{Dong2017BoostingAA}, and the hyperparameters are set as follows: the iterations are 100, $\lambda=1000$ in Eq.(\ref{eq:newobjective}), $\mu=1.0$ in Eq.(\ref{eq:mifgsm1}), $\epsilon=0.1$ in Eq.(\ref{eq:mifgsm2}).
	
	%comment: consider introducing the acronym MIFGSM
	
	SPADE~\cite{DBLP:conf/cvpr/Park0WZ19} is a method based on conditional normalization, and it can convert the segmentation map to a photo-realistic image. We use the SPADE in the silhouettes to video translation. %The training data setting is also the same as in Table~\ref{tab:setting}. 
	We train it on CASIA-B for 670000 iterations. %We report the experimental results under the attack setting as the 4th step in Fig.~\ref{fig:system}. When using the simple background difference algorithm in the signal processing, we obtain the same results for silhouettes as the input which we feed into the SPADE.
	
	%comment: the intent of this phrasing is unclear: we obtain the same silhouettes as the input which we feed into the SPADE.
	
	{\bf Settings for perturbation methods.}
	For comparison, we report the experimental results under the attack setting as the 4th step in Fig.~\ref{fig:system}. For perturbation-based methods, we choose FGSM, PGD, and MIFGSM as baselines. The distortion budget is set to the value 1 for these methods, with the pixel value within [0,1]. For iterative methods, PGD and MIFGSM, the iterations are set to 20. The decay factor in MIFGSM is set as 1.0.

\iffalse
\begin{figure*}
	\centering
	\includegraphics[width=1.0\linewidth]{fig4.pdf}
	%\captionsetup{font={footnotesize}}
	%\vspace{-5ex}
	\caption{One example of attacking on biometric samples. The first row shows the source sequence and the second row shows the corresponding adversarial sequence. The generated adversarial images are circled with red. It is not easy to notice the difference between the source sequence and our adversarial sequence in a real-time monitoring scene.}
	\label{fig:frames}
\end{figure*}
\fi

	\begin{figure*}
	\centering
	\subfigure[sparsity=1/40.]{
		\begin{minipage}[t]{0.32\linewidth}
			\centering
			\includegraphics[width=1\linewidth]{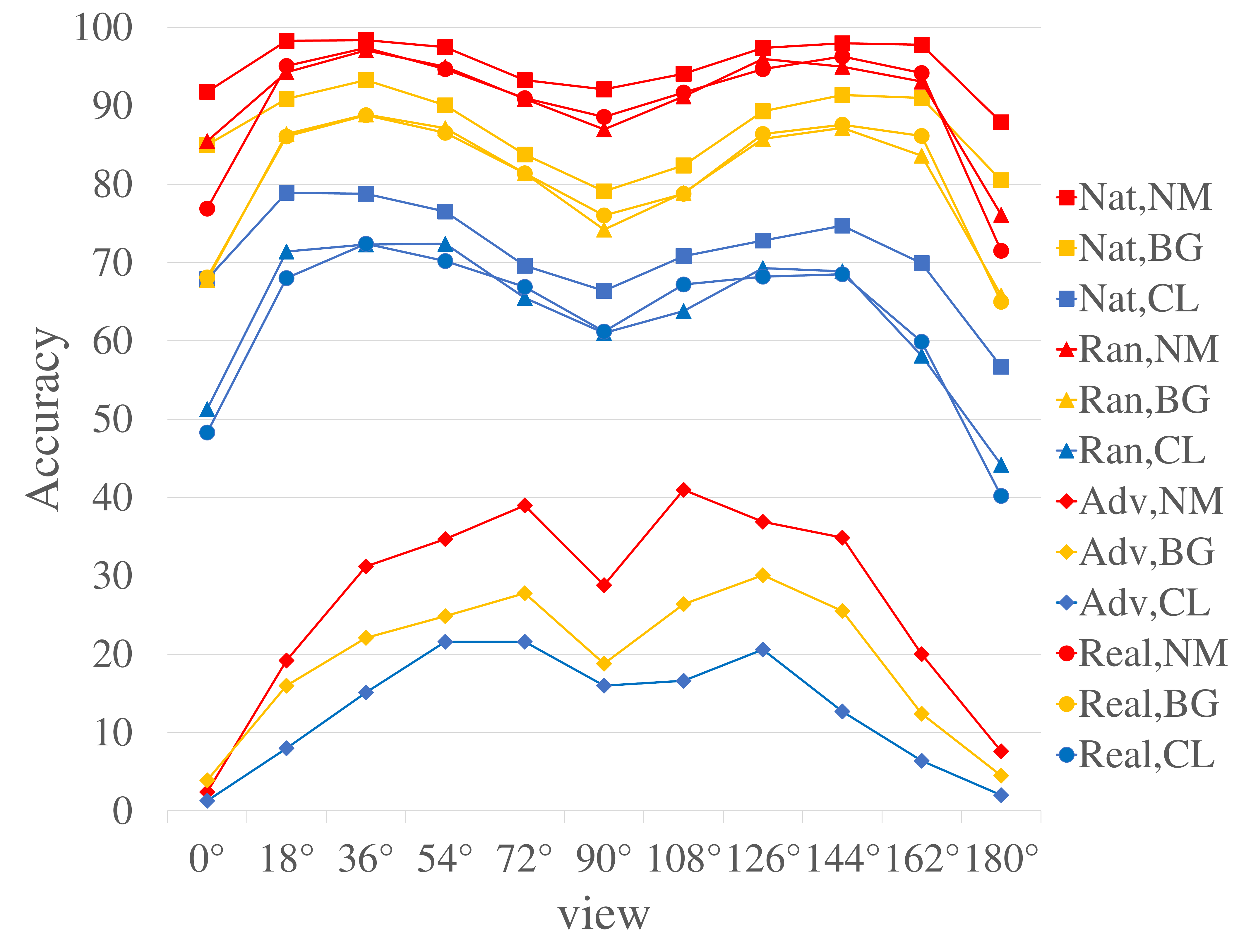}
			%\caption{fig1}
		\end{minipage}%
	}%
	\subfigure[sparsity=1/20.]{
		\begin{minipage}[t]{0.32\linewidth}
			\centering
			\includegraphics[width=1\linewidth]{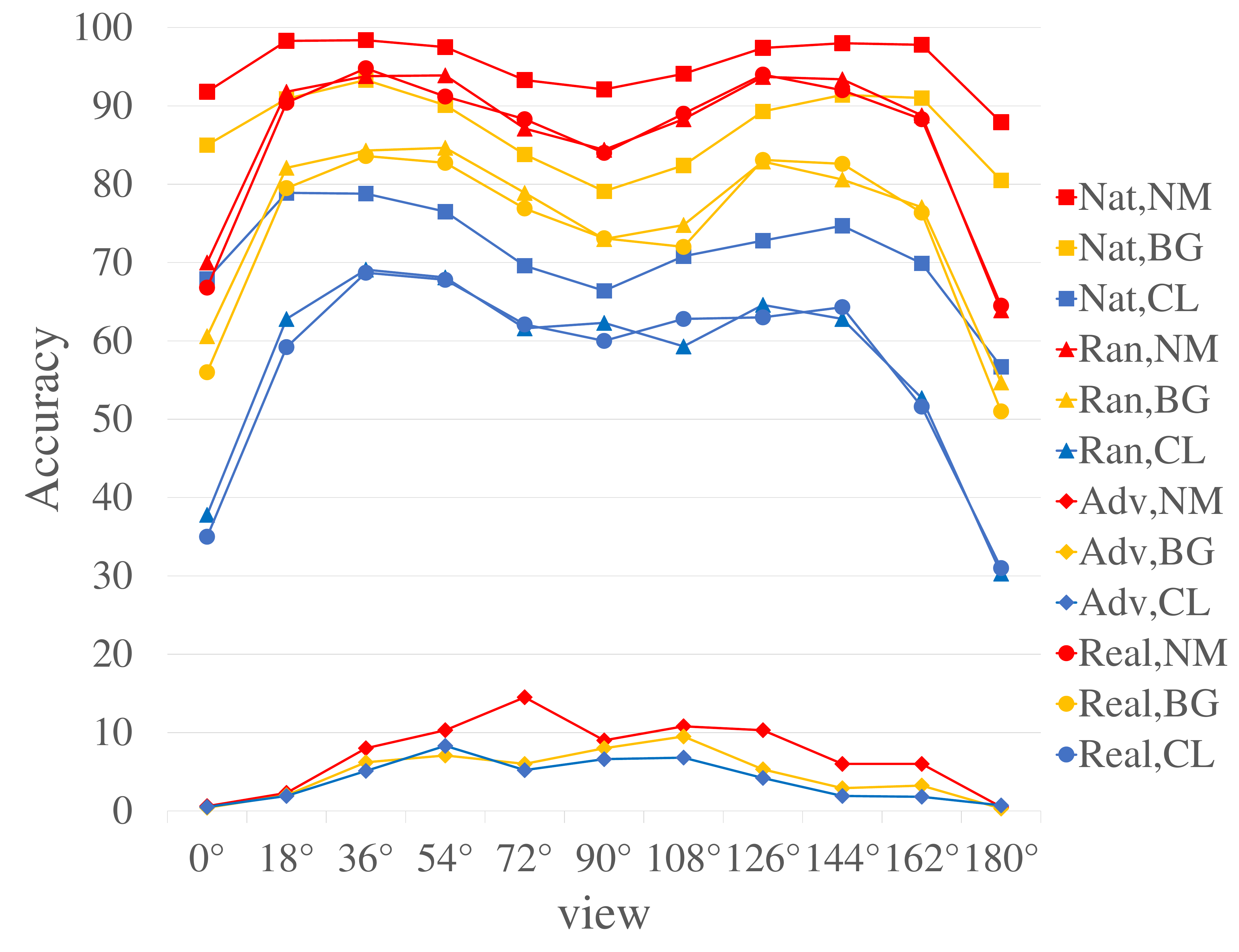}
			%\caption{fig2}
		\end{minipage}%
	}%
	\subfigure[sparsity=1/10.]{
		\begin{minipage}[t]{0.32\linewidth}
			\centering
			\includegraphics[width=1\linewidth]{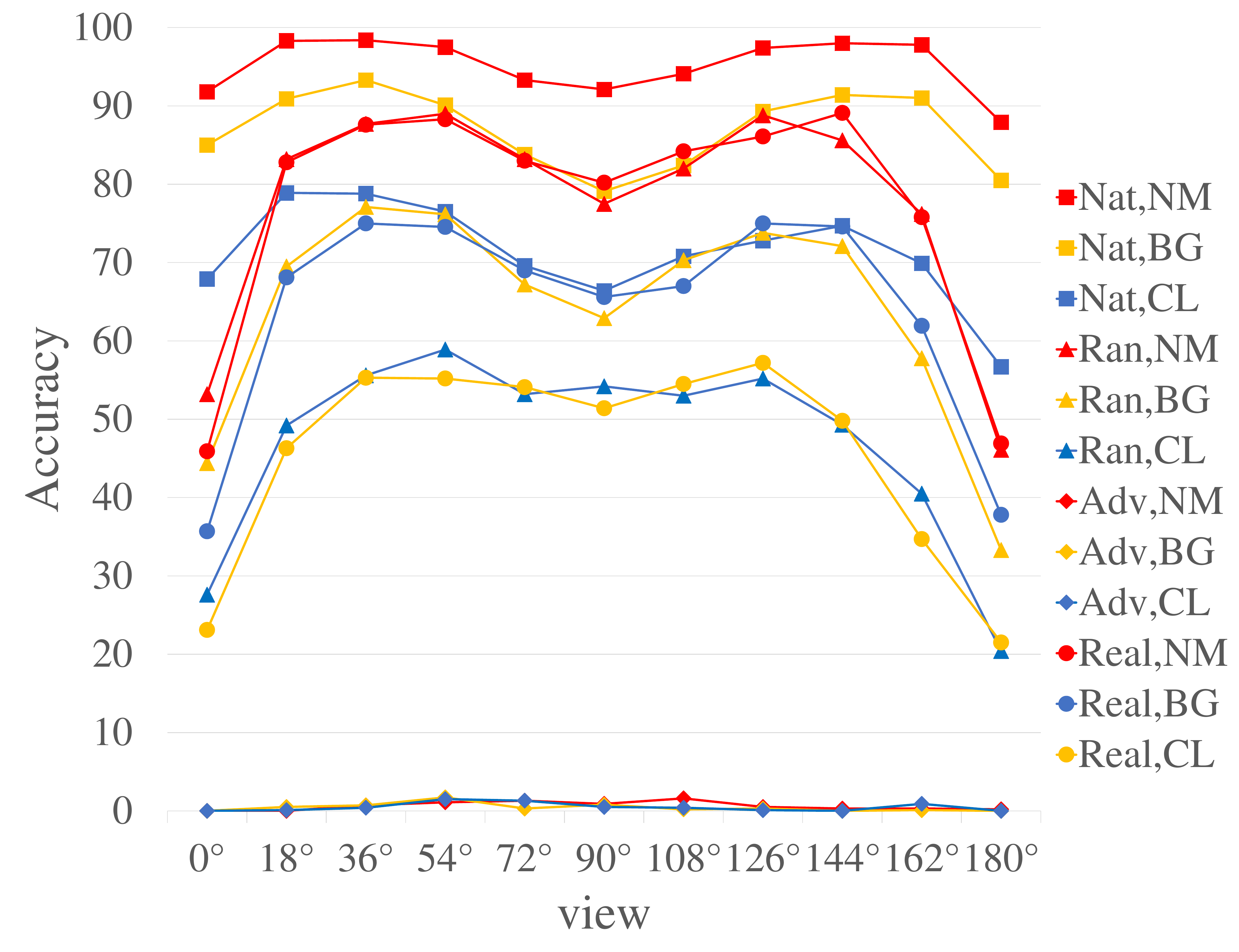}
			%\caption{fig2}
		\end{minipage}
	}%
	\centering
	\caption{Results of white-box nontargeted attack. Predefined sparsity includes 1/10, 1/20, and 1/40. Three walking conditions are labeled with different colors and four settings (Natural (Nat), Random (Ran), Real and Adversary (Adv)) are labeled with different legends.}
	\label{fig:results}
\end{figure*}

\begin{figure*}
	\centering
	\subfigure[NM.]{
		\begin{minipage}[t]{0.32\linewidth}
			\centering
			\includegraphics[width=1\linewidth]{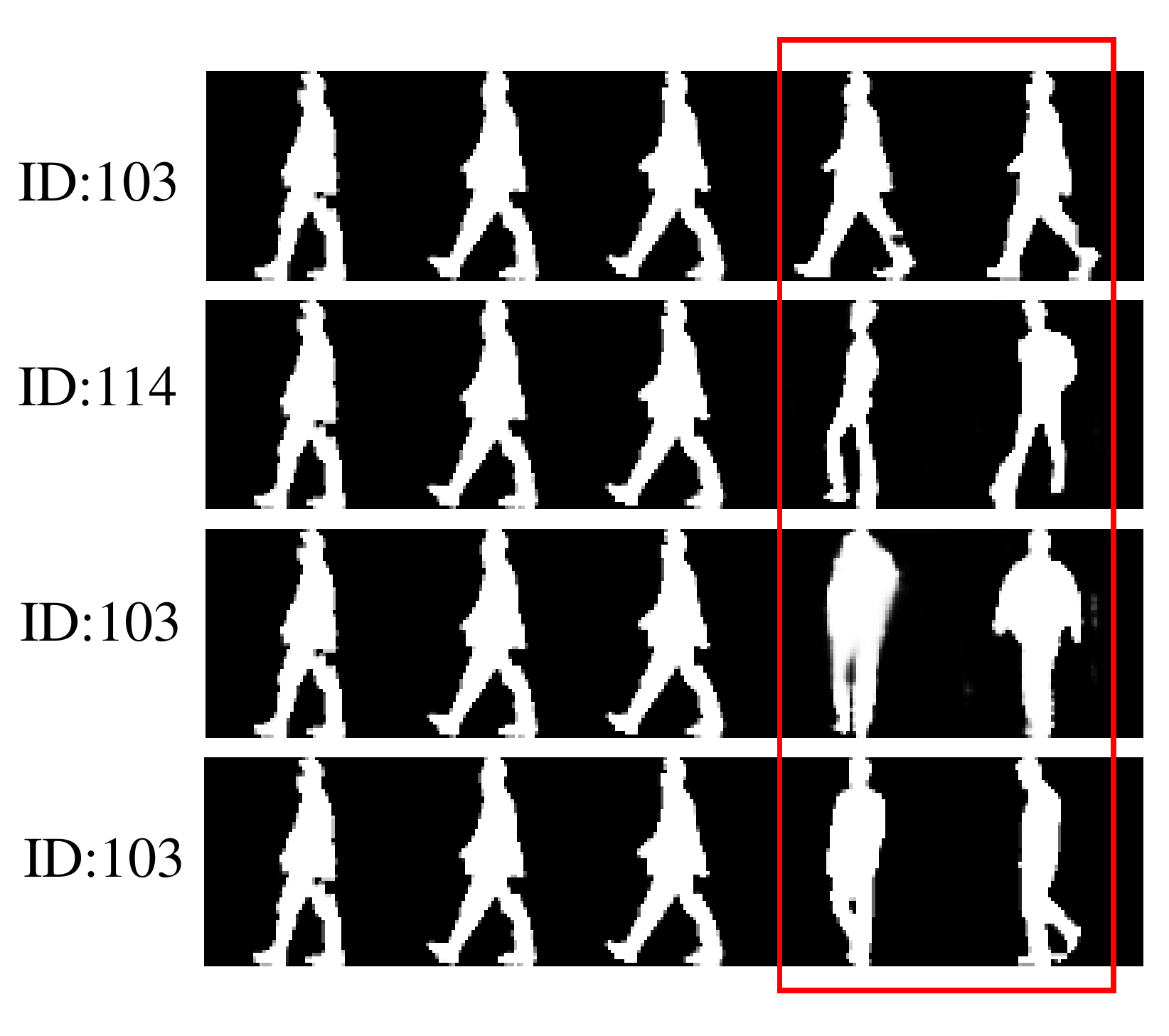}
			%\caption{fig1}
		\end{minipage}%
	}%
	\subfigure[BG.]{
		\begin{minipage}[t]{0.32\linewidth}
			\centering
			\includegraphics[width=1\linewidth]{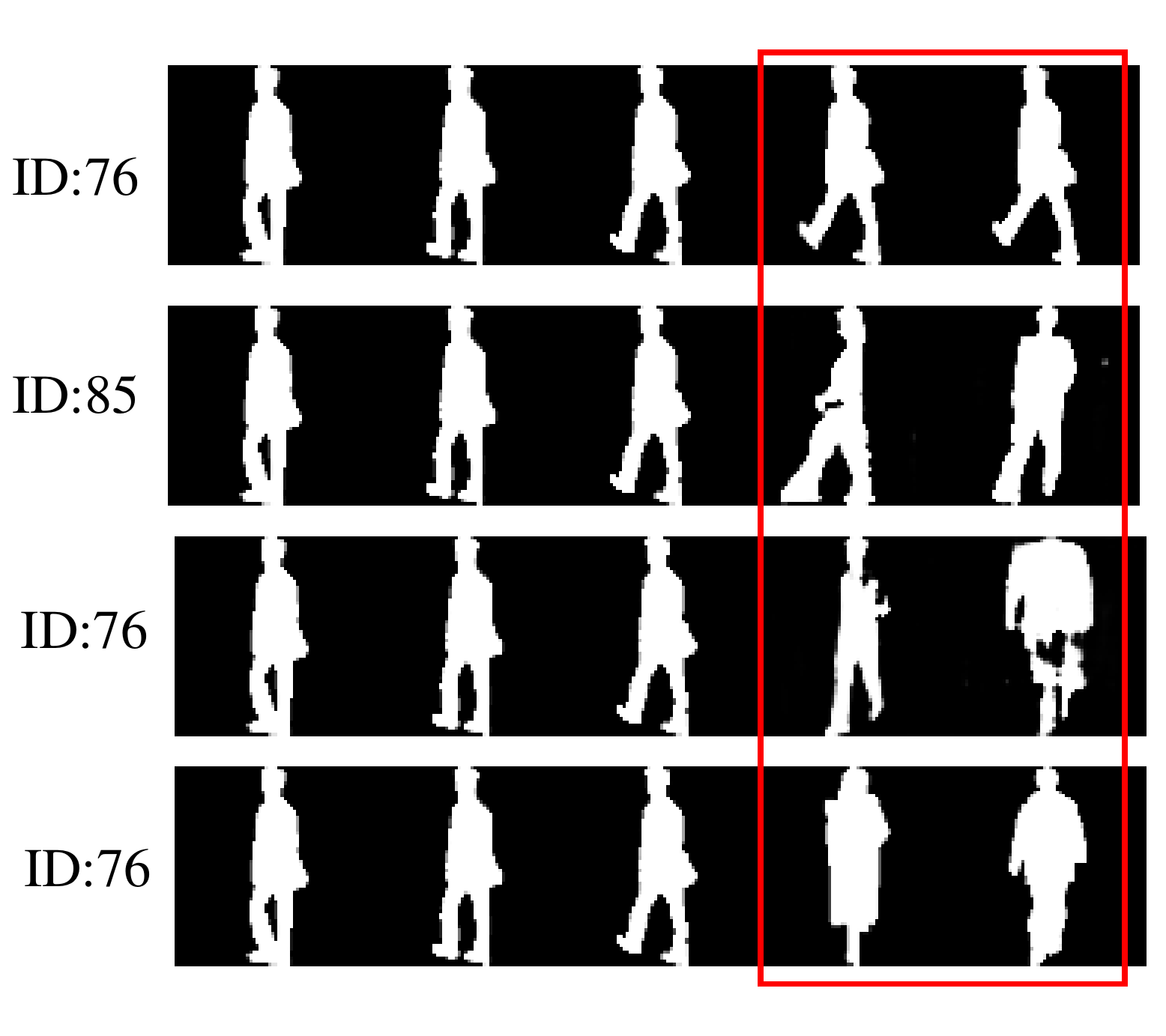}
			%\caption{fig2}
		\end{minipage}%
	}%
	\subfigure[CL.]{
		\begin{minipage}[t]{0.32\linewidth}
			\centering
			\includegraphics[width=1\linewidth]{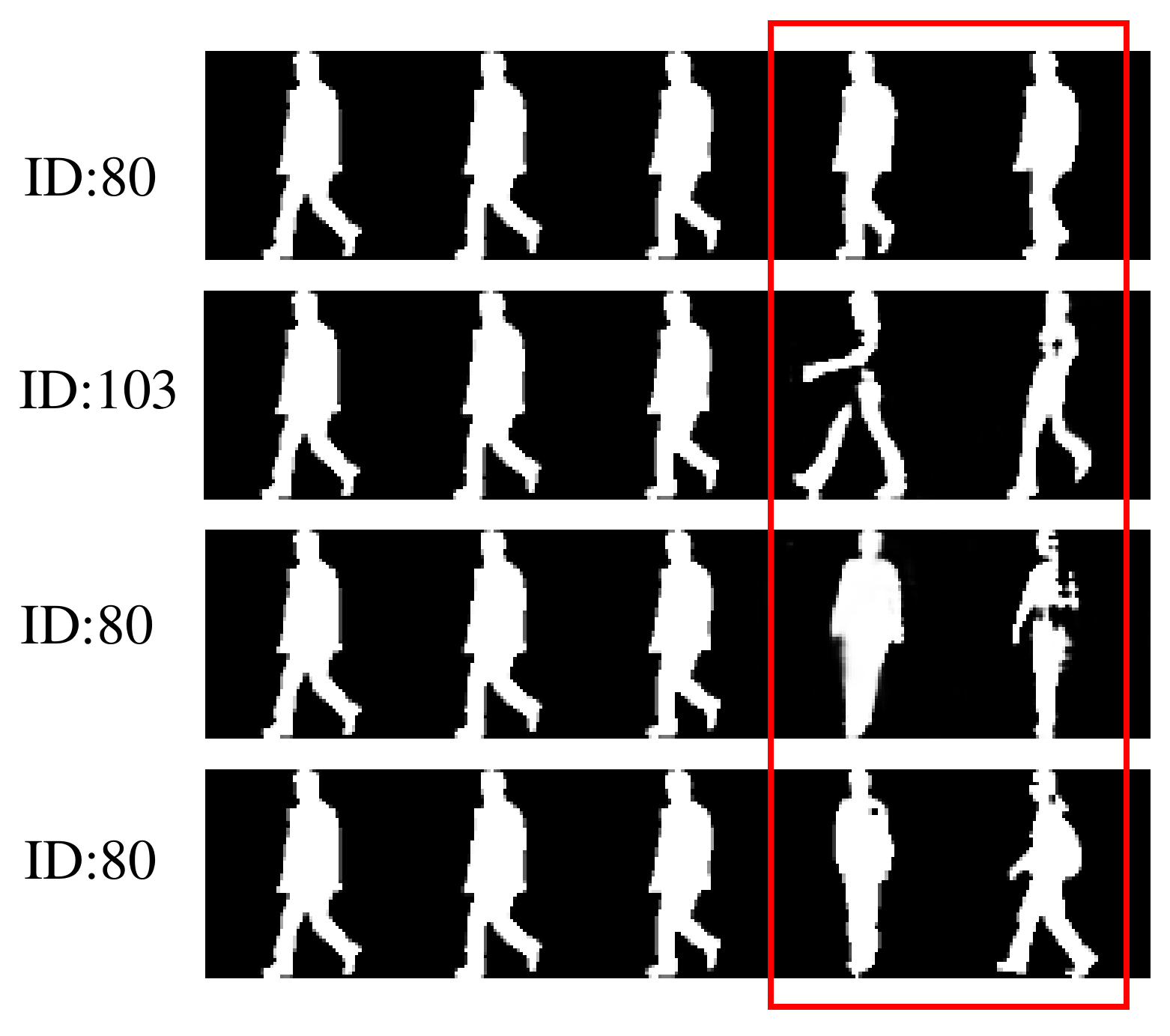}
			%\caption{fig2}
		\end{minipage}
	}%
	\centering
	\caption{Comparison of modified images. For each subfigure, from top to bottom are a short sequence corresponding to four different settings, i.e., \textit{Natural}, \textit{Adversarial}, \textit{Random} and \textit{Real}. ID is classified by GaitSet.}
	\label{fig:quality}
\end{figure*}
	
	\subsection{White-box experiments}
	\label{sec:results}
	
	In this subsection, we attack a gait model under the white-box protocol. This means we have the full knowledge of this target model. The attacked model is GaitSet, the state-of-the-art gait recognition model. GaitSet regards the gait as a set of gait silhouettes and utilizes a deep neural network to directly extract temporal information during training. Moreover, GaitSet is flexible since the input set can contain any number of non-consecutive silhouettes. 
	
	%For our attack on biometric samples, we list one adversarial sequence in Fig.~\ref{fig:frames}. The adversarial sequence shows satisfactory imperceptibility in temporal domain. Such a modification is hard to be detected in a real-time surveillance system. Moreover, in spatial domain, the adversarial images maintain the same background and contain subjects with a similar shape to real ones.
	
	In the following, we evaluate the attack ability of our method. For clarity, we report the results of frame-alteration attack and omit the results of frame-insertion attack as the trends for both methods are similar.
	%We report the results of our attack under two settings, nontargeted and targeted, in Fig.\ref{fig:results} and Table~\ref{tab:target}, respectively.
	%We report the results of our attack under two settings, frame-alteration and frame-insertion, in Fig.\ref{fig:results} and Fig.\ref{fig:perturbation}, respectively. The trends for both methods are similar.
	%The accuracy of GaitSet under our proposed attack is shown in Fig.\ref{fig:results}. 

	\textbf{Nontargeted results.}
	We report the results of our attack under nontargeted setting in Fig.\ref{fig:results}. The natural accuracy of GaitSet is labeled with \textit{Natural} and the accuracy under attack is labeled with \textit{Adversarial}. We observe that our attacks with different sparsity successfully deceive GaitSet, causing low accuracy in all three walking conditions. Moreover, the attack success rate is positively correlated with sparsity, which means that a stronger attack needs more modified frames in a sequence. However, the accuracy still drops dramatically when the sparsity is 1/40.  %Even though, the accuracy drops more than $68.2\%$ when the sparsity is 1/40. 

	\iffalse
	\begin{figure*}[t]
		\centering
		\includegraphics[width=1.0\linewidth]{fig4.pdf}
		%\captionsetup{font={footnotesize}}
		%\vspace{-5ex}
		\caption{Results of the frame-alteration attack. Predefined sparsity includes 1/10, 1/20, and 1/40. Three walking conditions are labeled with different colors and three settings (Natural (Nat), Random (Ran), and Adversary (Adv)) are labeled with different line types.}
		\label{fig:results}
	\end{figure*}
	\fi
	
	\iffalse
	\begin{figure*}
		\centering
		\subfigure[sparsity=1/40.]{
			\begin{minipage}[t]{0.33\linewidth}
				\centering
				\includegraphics[width=1\linewidth]{fig6-1.pdf}
				%\caption{fig1}
			\end{minipage}%
		}%
		\subfigure[sparsity=1/20.]{
			\begin{minipage}[t]{0.33\linewidth}
				\centering
				\includegraphics[width=1\linewidth]{fig6-2.pdf}
				%\caption{fig2}
			\end{minipage}%
		}%
		\subfigure[sparsity=1/10.]{
			\begin{minipage}[t]{0.33\linewidth}
				\centering
				\includegraphics[width=1\linewidth]{fig6-3.pdf}
				%\caption{fig2}
			\end{minipage}
		}%
		\centering
		\caption{Results of the \textbf{frame-insertion} attack. Predefined sparsity includes 1/10, 1/20, and 1/40. Three walking conditions are labeled with different colors and three settings (Natural (Nat), Random (Ran), and Adversary (Adv)) are labeled with different line types.}
		\label{fig:perturbation}
	\end{figure*}
	\fi
	
	To further prove the drop accuracy is caused by our novel attack design rather than altering some frames with randomly chosen gait silhouettes, we compare \textit{Adversarial} with other two situations: (1) \textit{Real}. Randomly selected real frames of other subjects replace some frames in the original sequence. (2) \textit{Random}. In this scenario, the latent variable $z$ is randomly sampled from a standard Gaussian distribution. Then our GAN model generates attacking frames from the randomly sampled vector $z$. This is different from our attacking method, since in our attack the vector $z$ is optimized by gradient backpropagation as the blue arrows in Fig.~\ref{fig:model}, i.e., our attack is searching the optimal $z$ in a prior distribution rather than randomly sampling $z$. As shown in Fig.~\ref{fig:results}, Both \textit{Real} and \textit{Random} only slightly affect the accuracy of GaitSet, while \textit{Adversarial} has more severe damage to the recognition performance. These results demonstrate the effectiveness of our attack. 
	
	One intriguing phenomenon is that altering some original frames have a more obvious effect on the accuracy when the view angle is close to $0^{\circ}$ or $180^{\circ}$. Under these conditions, the proposed attack is hardly recognizable. These remain challenging cases for most of the state-of-the-art gait recognition methods. Besides $0^{\circ}$ and $180^{\circ}$, the accuracy of $90^{\circ}$ under attack is a local minimum value. Chao et al.~\cite{Chao2018GaitSetRG} point out that both parallel and vertical perspectives lose some part of gait information. For example, stride can be observed most clearly at $90^{\circ}$ while a left-right swinging of body or arms can be observed most clearly at $0^{\circ}$ or $180^{\circ}$. For attacking case, we conclude that the parallel and vertical perspectives are more fragile when facing noises. Fig.~\ref{fig:results} empirically proves this statement. When replacing some frames with randomly generated or real silhouettes instead of adversarial images, the accuracy of $0^{\circ}$ or $180^{\circ}$ still has a larger decrease than other views.
	
	\begin{table*}
\caption{Results of white-box targeted attack, shown as attack success rate(\%). }
	\label{tab:target}
	\setlength{\tabcolsep}{3mm}
	\renewcommand{\arraystretch}{1.1}
	\begin{center}
		\begin{tabular}{|l|c|l|l|l|l|l|l|l|l|l|l|l|l|}
			\hline
			&    & \multicolumn{11}{c|}{viewing angles: 0$^{\circ}$ - 180$^{\circ}$}                                                            &         \\ \hline
			sparsity   &  condition & 0$^{\circ}$    & 18$^{\circ}$   & 36$^{\circ}$   & 54$^{\circ}$  & 72$^{\circ}$   & 90$^{\circ}$   & 108$^{\circ}$  & 126$^{\circ}$  & 144$^{\circ}$  & 162$^{\circ}$  & 180$^{\circ}$   & average \\ \hline\hline
			\multirow{3}{*}{1/40} & NM & 34.60 & 18.90 & 13.20 & 8.60  & 7.70  & 9.40  & 7.90  & 10.10 & 8.30  & 16.80 & 30.80 & 15.118  \\ \cline{2-14} 
			& BG & 31.70 & 20.70 & 14.60 & 11.41 & 11.70 & 14.40 & 12.40 & 12.10 & 16.10 & 22.93 & 35.50 & 18.504  \\ \cline{2-14} 
			& CL & 42.30 & 23.40 & 17.20 & 17.50 & 13.10 & 17.90 & 15.70 & 14.90 & 23.20 & 26.00 & 41.00 & 22.927  \\ \hline\hline
			\multirow{3}{*}{1/20} & NM & 61.60 & 43.40 & 33.30 & 30.50 & 25.40 & 29.00 & 22.20 & 27.90 & 37.00 & 40.10 & 63.50 & 37.627  \\ \cline{2-14} 
			& BG & 63.30 & 47.00 & 39.50 & 34.64 & 33.40 & 37.30 & 33.00 & 36.10 & 41.10 & 47.07 & 62.80 & 43.201  \\ \cline{2-14} 
			& CL & 65.60 & 49.20 & 39.50 & 37.80 & 35.70 & 36.20 & 36.30 & 38.60 & 47.20 & 50.70 & 63.80 & 45.509  \\ \hline\hline
			\multirow{3}{*}{1/10} & NM & 78.60 & 67.40 & 59.40 & 61.40 & 60.00 & 61.10 & 58.90 & 60.90 & 65.10 & 68.60 & 80.60 & 65.636  \\ \cline{2-14} 
			& BG & 82.30 & 69.80 & 65.40 & 64.55 & 61.20 & 62.90 & 60.50 & 63.50 & 66.50 & 70.91 & 80.20 & 67.978  \\ \cline{2-14} 
			& CL & 78.90 & 70.50 & 63.80 & 66.20 & 62.40 & 65.00 & 65.50 & 65.70 & 69.90 & 73.10 & 81.30 & 69.300  \\ \hline
		\end{tabular}
	\end{center}
\end{table*}

\begin{table*}
	\caption{Comparison with spoofing attack, shown as success rate(\%). }
	\label{tab:3}
	\begin{center}
		%\resizebox{85mm}{10mm}{
		\setlength{\tabcolsep}{3mm}
		\renewcommand{\arraystretch}{1.1}
		\begin{tabular}{|l|l|l|l|l|l|l|l|l|l|l|l|l|l|}
			\hline
			gallery: NM 01-04 & \multicolumn{13}{c|}{viewing angles: 0$^{\circ}$ - 180$^{\circ}$} \\ \hline
			probe: NM 05-06   & 0$^{\circ}$    & 18$^{\circ}$   & 36$^{\circ}$   & 54$^{\circ}$  & 72$^{\circ}$   & 90$^{\circ}$   & 108$^{\circ}$  & 126$^{\circ}$  & 144$^{\circ}$  & 162$^{\circ}$  & 180$^{\circ}$  & average & sparsity\\ \hline
			spoofing attack~\cite{Jia2019AttackingGR} & 68.0& 86.0& 92.0& 89.0& 82.0& 78.0& 82.0 &89.0& 90.0 &85.0& 65.0 &82.0&100$\%$\\ \hline
			%ours & 78.6& 67.4& 59.4& 61.4& 60.0& 61.1& 58.9&60.9& 65.1& 68.6& 80.6& 65.6 &10$\%$\\
			ours & 85.3& 80.1& 75.3& 74.2& 74.7& 76.9& 76.6& 75.4& 77.3& 78.2& 84.0&78.0&20$\%$\\
			\hline
		\end{tabular}
	\end{center}
\end{table*}

	\begin{table*}
	\caption{Results of a Black-box attack on GaitGAN, shown as accuracy(\%). }
	\label{tab:black}
	\begin{center}
		%\resizebox{85mm}{10mm}{
		\setlength{\tabcolsep}{3.9mm}
		\renewcommand{\arraystretch}{1.1}
		\begin{tabular}{|l|l|l|l|l|l|l|l|l|l|l|l|l|}
			\hline
			probe view   & 0$^{\circ}$    & 18$^{\circ}$   & 36$^{\circ}$   & 54$^{\circ}$  & 72$^{\circ}$   & 90$^{\circ}$   & 108$^{\circ}$  & 126$^{\circ}$  & 144$^{\circ}$  & 162$^{\circ}$  & 180$^{\circ}$  & average\\ \hline
			natural      & 39.4 & 56.0 & 62.3 & 61.1 & 59.3 & 25.8 & 55.8 & 63.6 & 57.3 & 52.9 & 40.7& 52.2 \\ \hline
			after attack & 35.9 & 52.8 & 60.0 & 57.6 & 56.1 & 24.4 & 52.2 & 60.4 & 55.2 & 49.5 & 36.2 & 49.1\\ \hline
			drop $\downarrow$       & 3.5  & 3.2  & 2.3  & 3.5  & 3.2  & 1.4  & 3.6  & 3.2  & 2.1  & 3.4  & 4.5 & 3.1 \\ \hline
		\end{tabular}
	\end{center}
\end{table*}
	
	Moreover, to prove that the accuracy is not affected by the quality of inserted images, we randomly select some silhouettes and show them in Fig.~\ref{fig:quality}. Here the source sequence of NM is five consecutive frames in the fifth sequence of subject 103 under normal condition with angle view $108^{\circ}$, BG is the second sequence of subject 76 with angle view $144^{\circ}$ and CL is the first sequence of subject 80 with angle view $72^{\circ}$. For all the three walking conditions, NM, BG, and CL, the silhouettes drawn from attacking frames achieve a competitive quality with a real silhouette. Furthermore, we observe that, although the generated silhouettes of \textit{Random} in Fig.~\ref{fig:quality}(b) are low quality and do not seem like a person with bag, GaitSet still makes a correct classification. Contrarily, the adversarial sequence successfully fools GaitSet. Therefore, GaitSet is robust to a slight disturbance but vulnerable under the proposed adversarial attack.

	\textbf{Targeted results.}
	We report the results of targeted attack in Table~\ref{tab:target}. The targeted results share some similarity with nontargeted results: the attack success rate is positively correlated with sparsity; the attack achieves a higher success rate when the view angle is $0^{\circ}$ or $180^{\circ}$. But compared with nontargeted one, targeted attack is apparently more difficult, because it aims to deceive the gait recognition system with a specific subject ID rather than any one. Nonetheless, the proposed method can successfully deceive the GaitSet with a high rate at around 65$\%$ when the sparsity is 1/10. Therefore, our method can serve as a strong benchmark of adversarial attack on gait recognition. 

	The goal of targeted attack is similar to spoofing attack, which aims to gain illegitimate access to gait systems by masquerading as others. Here we compare our targeted attack with spoofing attack proposed by Jia et al~\cite{Jia2019AttackingGR}. Results are shown in Table~\ref{tab:3}. Though spoofing attack achieves a high success rate, it needs to alter each frame of the source sequence, i.e., generating a fake background to substitute the original background. Our method can achieve a satisfactory fooling rate with a slighter modification.

\begin{figure*}
	\centering
	\subfigure[sparsity=1/40.]{
		\begin{minipage}[t]{0.3\linewidth}
			\centering
			\includegraphics[width=1\linewidth]{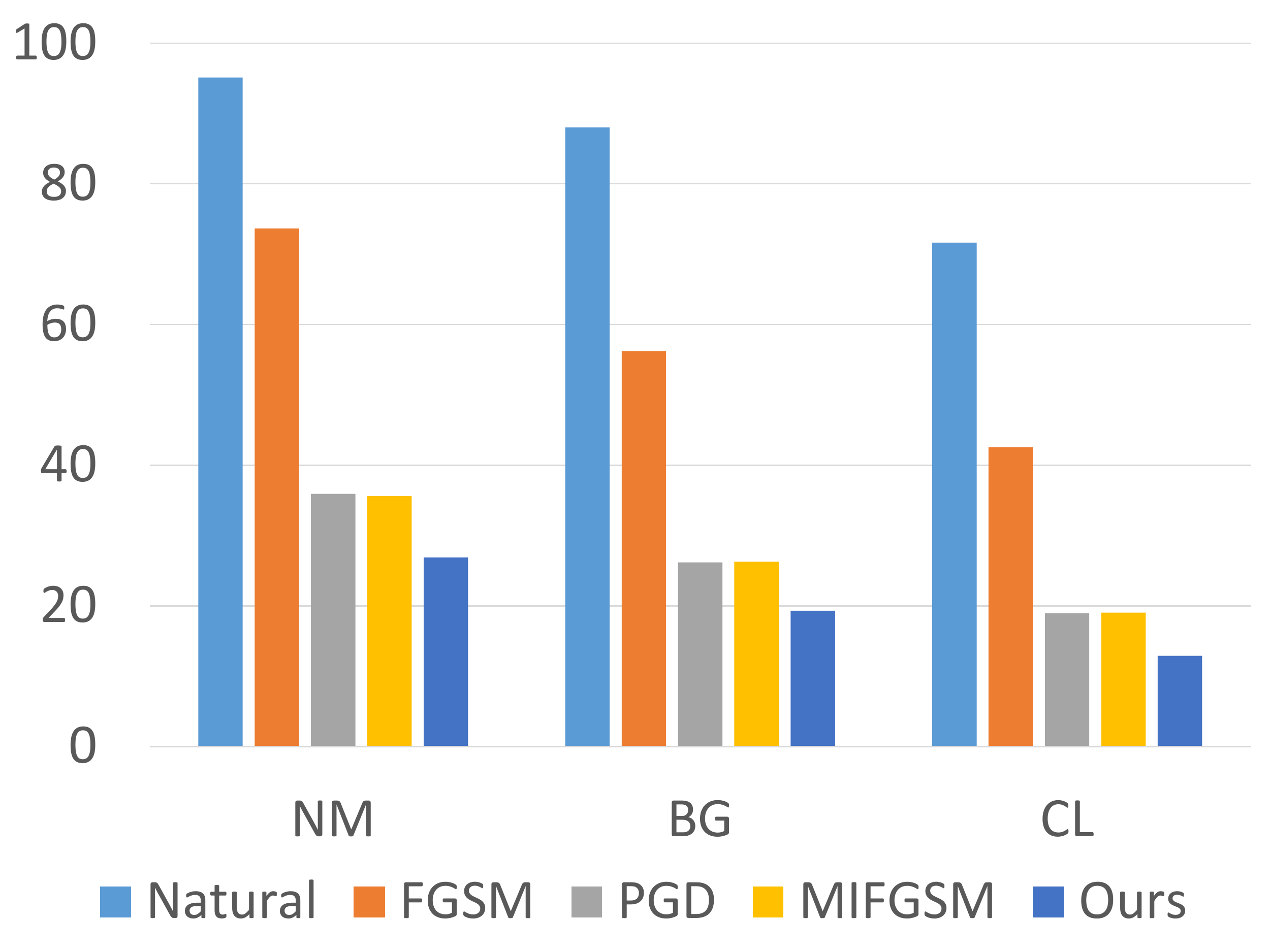}
			%\caption{fig1}
		\end{minipage}%
	}%
	\subfigure[sparsity=1/20.]{
		\begin{minipage}[t]{0.3\linewidth}
			\centering
			\includegraphics[width=1\linewidth]{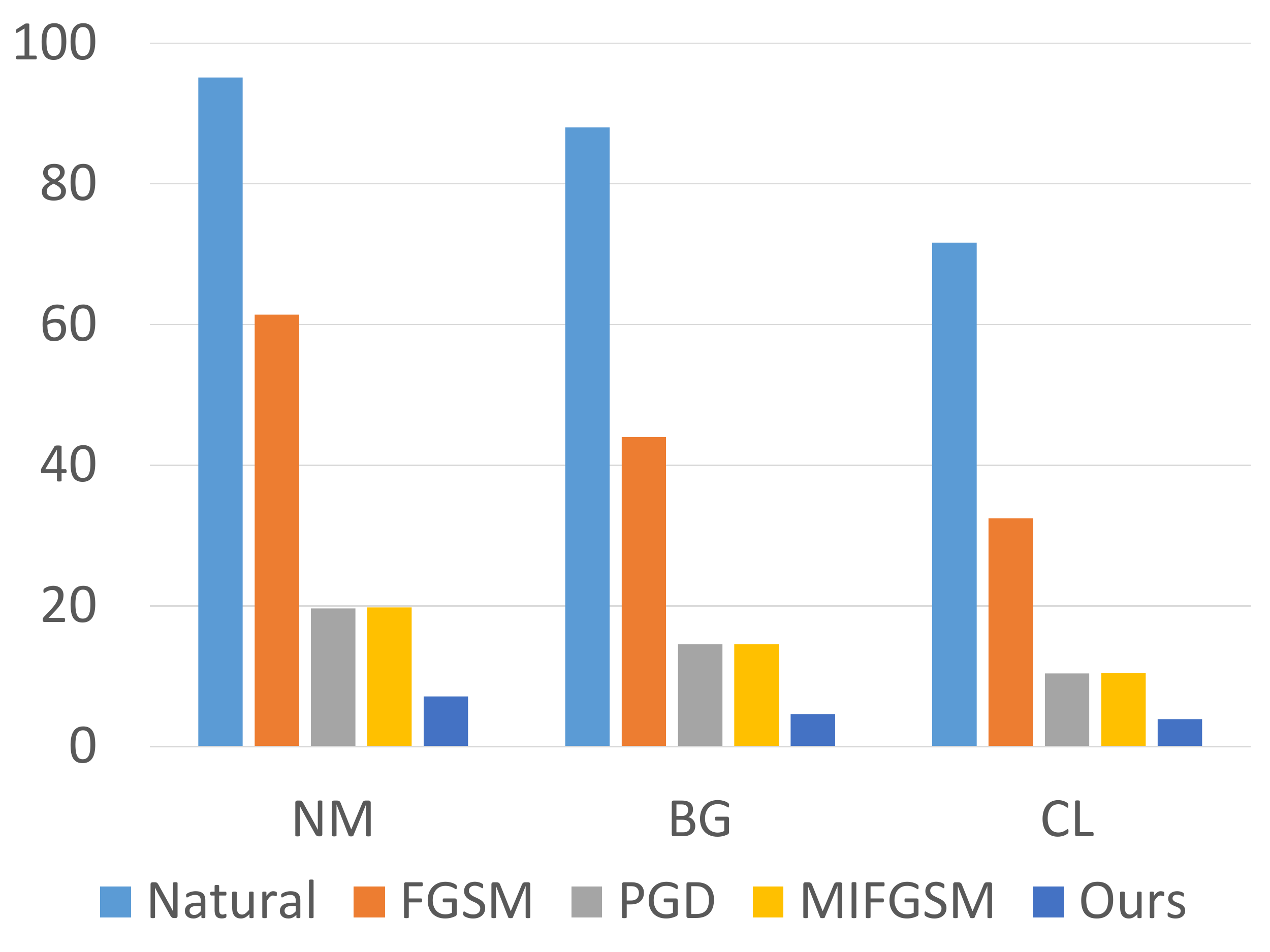}
			%\caption{fig2}
		\end{minipage}%
	}%
	\subfigure[sparsity=1/10.]{
		\begin{minipage}[t]{0.3\linewidth}
			\centering
			\includegraphics[width=1\linewidth]{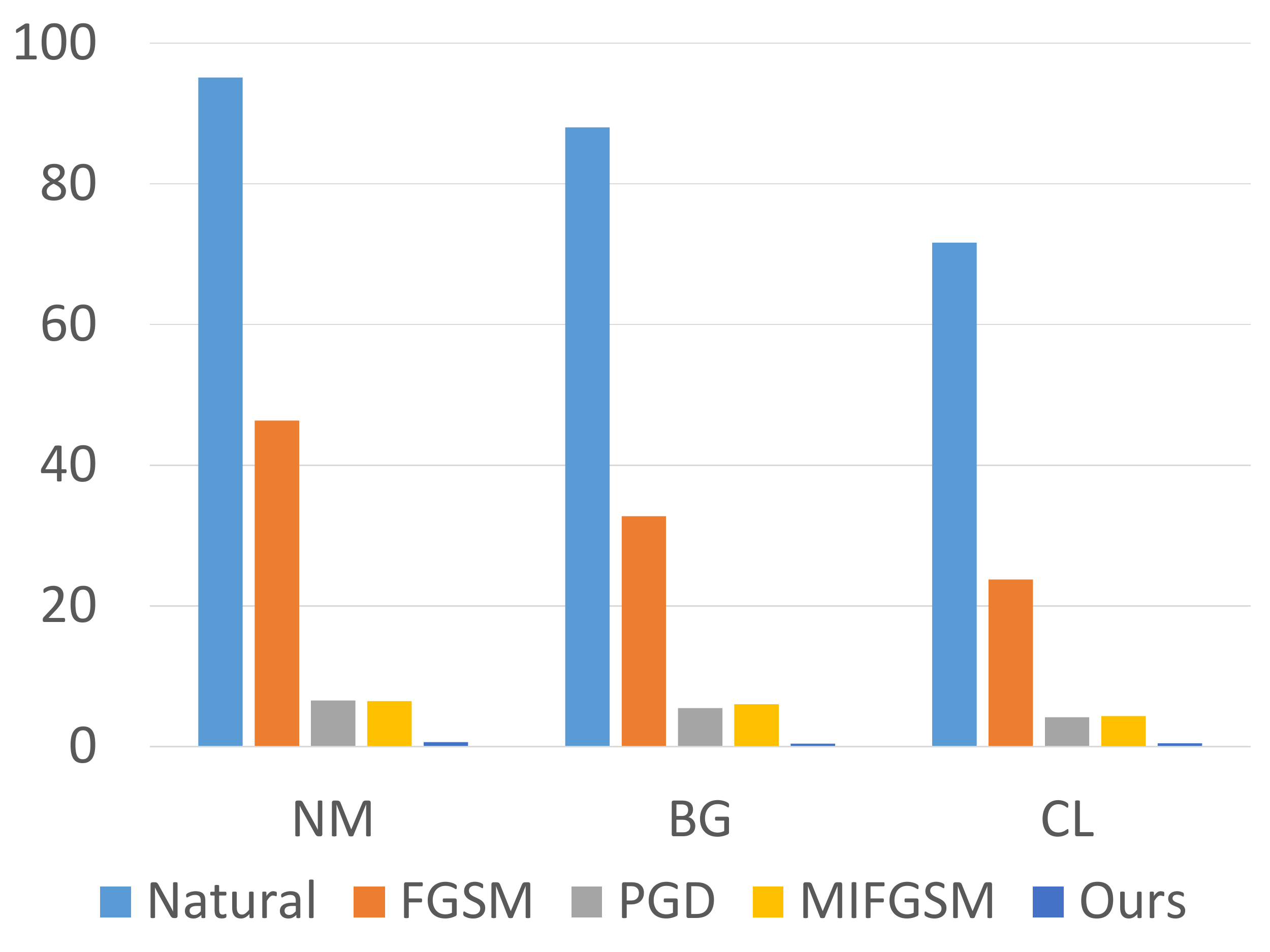}
			%\caption{fig2}
		\end{minipage}
	}%
	\centering
	\caption{Comparison of our method with perturbation-based methods. `Natural' is the original accuracy of GaitSet and others represent the accuracy under attack.}
	\label{fig:7}
\end{figure*}
	
	\subsection{Black-box experiments}
	\label{sec:black}
	Our attack method is specifically aimed at sequence-based gait recognition models, and the above experimental results demonstrate their vulnerability. In this subsection, we also make a black-box attack on the template-based model, GaitGAN~\cite{DBLP:conf/cvpr/YuCRP17}. Different from GaitSet, which takes a gait sequence as a set and extracts its feature with a CNN, GaitGAN uses a GEI template as the gait feature. Moreover, GaitGAN takes a GAN model as a regressor to simultaneously address variations in viewpoint, clothing, and carrying conditions in gait recognition. In the black-box scenario, we cannot access any information of GaitGAN in the attack process. To perform the black-box attack, we apply the widely used transfer-based attack~\cite{Papernot2016TransferabilityIM,Liu2016DelvingIT}. It leverages a property of adversarial examples, i.e., transferability, which means that adversarial examples crafted on one model can successfully attack another model with different architecture and parameters. In transfer-based attack, attackers use a local substitute model to craft adversarial examples and feed them into a black-box target model to result in wrong outputs. Specifically, here we firstly attack GaitSet with Algorithm~\ref{alg:attack} to obtain the adversarial sequence, and then use it as the input of GaitGAN to test the accuracy.
	
	%The experimental setting is the same as CASIA-B in Table~\ref{tab:setting}. 
	
	The sparsity is $1/40$ and the adversarial sequences are the same as Adv, NM in Fig.~\ref{fig:results}(a). We report the results of probeNM, shown in Table~\ref{tab:black}. The recognition rate of each probe view only drops a little after attacking. We conclude that GEI is more robust than the feature extracted by GaitSet under our temporal adversarial attack. Because GEI is obtained by aligning the silhouettes in the spatial space and averaging them along the temporal dimension, the perturbation of a few frames is not enough to deceive GaitGAN. Although sequence-based gait recognition has made great progress in recognition accuracy, its robustness compared to template-based methods remains limited. This is a key area for the community to focus on in the future.
	
	%comment: consider clarifying "performance"

	\iffalse
	\begin{table}[t]
		\caption{Black-box attack on GaitGAN. Results are shown as accuracy(\%), `avg' is the average of all views.}
		\label{tab:black}
		\begin{center}
			%\resizebox{85mm}{10mm}{
			\begin{tabular}{|l|l|l|l|l|l|l|}
				\hline
				probe view   & 0    & 18   & 36   & 54   & 72   & 90   \\ \hline
				natural      & 39.4 & 56.0 & 62.3 & 61.1 & 59.3 & 25.8 \\ \hline
				after attack & 35.9 & 52.8 & 60.0 & 57.6 & 56.1 & 24.4  \\ \hline
				drop $\downarrow$       & 3.5  & 3.2  & 2.3  & 3.5  & 3.2  & 1.4  \\ \hline \hline
				probe view   & 108  & 126  & 144  & 162  & 180 & avg\\ \hline
				natural      & 55.8 & 63.6 & 57.3 & 52.9 & 40.7 & 52.2\\ \hline
				after attack & 52.2 & 60.4 & 55.2 & 49.5 & 36.2 & 49.1\\ \hline
				drop $\downarrow$  & 3.6  & 3.2  & 2.1  & 3.4  & 4.5 & 3.1\\ \hline
			\end{tabular}
			%}
		\end{center}
	\end{table}
	\fi

	\subsection{Cross-dataset validation}
	\label{sec:cross}
	For a more reliable performance assessment, we conduct cross-database testing using CASIA-A. In this scenario, the training set of CASIA-B is used to train the GaitSet and WGAN-GP, while the whole CASIA-A dataset is used for testing. %Detail settings are in Table~\ref{tab:setting}. 
	Results are shown in Table~\ref{tab:cross}. The accuracy is averaged on the $3$ gallery views, and the identical views are included. The trend is almost the same as the results of testing on CASIA-B. The recognition capability of the attacked model drops rapidly as the attack sparsity increases. 
	
	\begin{table}
		\caption{Results of cross-database validation with nontargeted attack, shown as accuracy(\%).}
		\label{tab:cross}
		\setlength{\tabcolsep}{4mm}
		\renewcommand{\arraystretch}{1.1}
		\begin{center}
			\begin{tabular}{|l|l|l|l|l|}
				\hline
				& \multicolumn{4}{c|}{probe view}\\
				\hline
				sparsity & $0^{\circ}$     & $45^{\circ}$    & $90^{\circ}$    & average \\ \hline
				0        & 56.67 & 70.00 & 76.67 & 67.78   \\ \hline
				1/40     & 33.33 & 33.33 & 18.33 & 28.33   \\ \hline
				1/20     & 18.33 & 25.00 & 3.33  & 15.55   \\ \hline
				1/10     & 6.67  & 8.33  & 3.33  & 5.11    \\ \hline
			\end{tabular}
		\end{center}
	\end{table}
	
	Under nontargeted setting, the performance degradation could be affected by many reasons, such as domain shift or the generalization ability of the recognition method itself, other than attacking. For a more convincing justification, we further perform cross-database targeted attack. Results are shown in Table~\ref{tab:cross_target}. When the sparsity is 1/10, the attack success rate reaches $68.33\%$ on average.
	
	\begin{table}
		\caption{Results of cross-database validation with targeted attack, shown as targeted attack success rate(\%).}
		\label{tab:cross_target}
		\setlength{\tabcolsep}{4mm}
		\renewcommand{\arraystretch}{1.1}
		\begin{center}
			\begin{tabular}{|l|l|l|l|l|}
				\hline
				& \multicolumn{4}{c|}{probe view}\\
				\hline
				sparsity & $0^{\circ}$     & $45^{\circ}$    & $90^{\circ}$    & average \\ \hline
				%0        & 56.67 & 70.00 & 76.67 & 67.78   \\ \hline
				1/40     & 11.67 & 8.33 & 31.67 & 17.22   \\ \hline
				1/20     & 26.67 & 20.00 & 61.67  & 36.11   \\ \hline
				1/10     & 56.67  & 63.33  & 85.00  & 68.33    \\ \hline
			\end{tabular}
		\end{center}
	\end{table}

	\subsection{Comparison with perturbation-based methods}
	\label{sec:compare}
	In Sec.~\ref{sec:intro}, we have qualitatively demonstrated the shortcomings of the perturbation-based approaches. In this subsection, we compare our proposed method with these methods quantitatively. The perturbation-based attacks used as baselines include FGSM~\cite{Goodfellow2014ExplainingAH},  PGD~\cite{Madry2017TowardsDL} and MIFGSM~\cite{Dong2017BoostingAA}.  %First, the perturbation is notable, compared with the proposed temporal sparse adversarial attack. Moreover, 
	It is difficult to extend these attacks to video frame generation due to the signal processing in silhouette-based gait recognition. Therefore, in this study we perform attacks on the gait silhouettes. For a fair comparison, we fix the sparsity; the attack success rate and imperceptibility of adversarial examples generated by different methods are compared. The distortion budgets of perturbation-based methods are all relaxed to pixel values, which means adversarial examples are not norm-bounded by a small constant for these methods. Our method is under the protocol of a frame-alteration attack.

\begin{figure}
		\centering
		\includegraphics[width=0.8\linewidth]{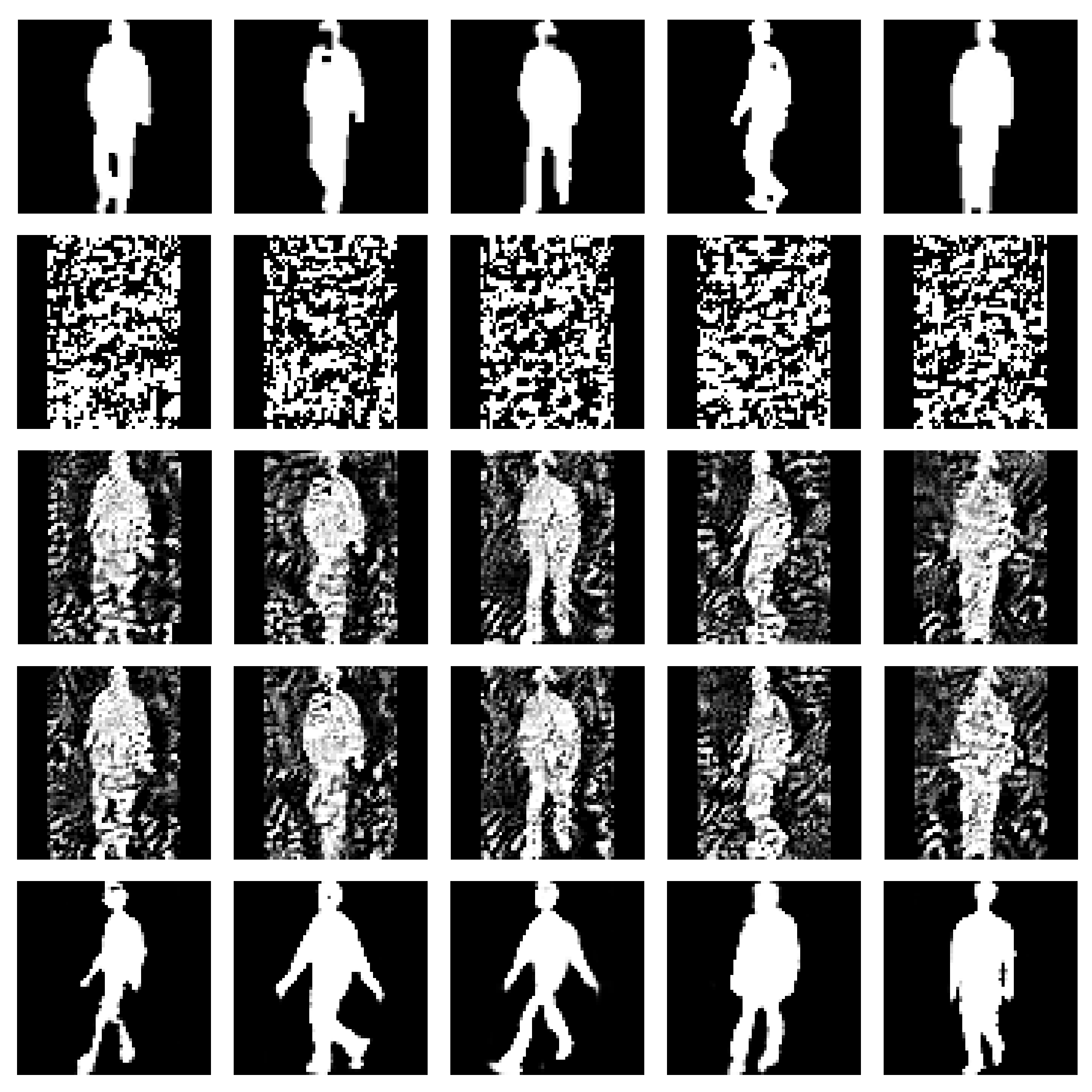}
		\caption{From top to bottom are natural examples and their corresponding adversarial examples generated by FGSM, PGD, MIFGSM, and our method.}
		\label{fig:8}
	\end{figure}

	The results are shown in Fig.~\ref{fig:7}, and some crafted adversarial examples are shown in Fig.~\ref{fig:8}. Our proposed method achieves a superior attack success rate and imperceptibility. In Fig.~\ref{fig:7}, we observe that while all of the attacks lower the accuracy of GaitSet, our method surpasses the perturbation-based methods by obtaining superior attack success rates in all of the settings. In Fig.~\ref{fig:8}, we show that the subjects in the silhouettes retain a human posture in our method. Thus, it maintains better imperceptibility of the spatial domain than perturbation-based methods. Furthermore, it enables the transfer of these silhouettes to video frames; it makes a practical threat to the gait recognition system. However, a limitation is that the generated samples have pose changes when they are compared to their adjacent frames. Therefore, some constraints are needed to enforce the changes between the adjacent frames, which we leave for our future work. Compared with perturbation-based methods, our proposed method provides superior success rates and imperceptibility and can serve as a stronger baseline for sequence-based gait recognition.
	
	%comment: consider clarifying "performance"
	\subsection{Analysis of the proposed method}
	\label{sec:analysis}
	In this section, we make a further analysis of the proposed method.
	
	\textbf{Position of frame-insertion.} Firstly, we study the effects of position to insert the adversarial image. In our prior experiments, we use GaitSet as the target model. GaitSet has achieved state-of-the-art performances without modeling the temporal characteristics explicitly. In other words, GaitSet takes a set of silhouettes as input and the order of input frames does not affect the recognition. Similarly, models using gait templates, such as GaitGAN~\cite{DBLP:conf/cvpr/YuCRP17}, aggregate temporal walking information over a sequence of silhouettes in a single map. The order of a gait sequence does not matter in these methods. Differently, some models learn from the order and relationship of frames in gait sequences, instead of aggregating them. We take SelfGait~\cite{liu2021selfgait} as an example and perform nontargeted attack on it. SelfGait is a self-supervised framework with spatiotemporal components to learn from the massive unlabeled gait images. Since SelfGait preserves and learns temporal representation from the order and relationship of frames in gait sequences, disrupting the order of input frames will decrease its accuracy. 
	
	In the following experiments, we randomly draw 30 sorted successive frames from a sequence as input and insert 4 adversarial frames into different positions. Results are reported in Table~\ref{tab:position}. Different positions are numbered with the index of adversarial frames in the obtained 34 frame-length sequence. For example, $\{0,1,2,3\}$ represents that all the 4 adversarial frames are inserted into the start of a sequence. In Table~\ref{tab:position}, from the row of $\{0,1,2,3\}$ to $\{30,31,32,33\}$ are inserting all the 4 frames into two adjacent frames. From the row of $\{0,11,22,33\}$ to $\{13,15,17,19\}$ are inserting adversarial frames into equidistant positions. The last row, `random', means that for each original sequence, adversarial frames are inserted into randomly chosen positions. Its final result is averaged on five experiments with different random seeds (from 0 to 4). We observe different positions have a slight effect on the attacking performance. For example, for all inserting positions, the accuracy of SelfGait under condition NM is around 19$\%$. Therefore, our method can pre-define any positions to insert the generated adversarial frames for a similar result. 
	
	\begin{table}
	\caption{Study on position of inserted adversarial frames, shown as accuracy(\%). }
	\label{tab:position}
	\setlength{\tabcolsep}{2mm}
	\renewcommand{\arraystretch}{1.1}
	\begin{center}
		\begin{tabular}{|c|c|c|c|}
			\hline
			& \multicolumn{3}{c|}{walking condition} \\ \hline
			position & NM          & BG          & CL         \\ \hline
			no inserted frames     & 91.079      & 84.735      & 74.261    \\ \hline
			$\{$0,1,2,3$\}$    & 17.711      & 12.197      & 10.663     \\ \hline
			$\{$8,9,10,11$\}$     & 18.895      & 11.193      & 11.212     \\ \hline
			$\{$15,16,17,18$\}$     & 18.190      & 11.913      & 11.629     \\ \hline
			$\{$23,24,25,26$\}$     & 18.176      & 11.420      & 12.689     \\ \hline
			$\{$30,31,32,33$\}$     & 19.784      &  12.651      & 11.117     \\ \hline
			$\{$0,11,22,33$\}$     & 19.607      & 12.784      & 11.723     \\ \hline
			$\{$4,12,20,28$\}$     & 20.035      & 14.565      & 11.553     \\ \hline
			$\{$9,14,19,24$\}$     & 18.406      & 11.572      & 9.583     \\ \hline
			$\{$13,15,17,19$\}$     & 18.992      & 11.553      & 10.303     \\ \hline
			random (mean$\pm$std)     & 19.130$\pm$0.406      & 12.940$\pm$0.281      & 10.530$\pm$0.890     \\ \hline
			
		\end{tabular}
	\end{center}
	
\end{table}
	
	\textbf{Number of adversarial images.} We study the minimum number of inserted adversarial frames in order to yield a satisfactory result. To make a fair comparison, we fix the length of silhouette frames in the test phase. Specifically, the sampler collects 30 sorted successive frames as input. We evaluate under two settings: (1) Static evaluation. We pre-define a number of inserted adversarial frames and calculate the fooling rate under such a pre-defined number. (2) Dynamic evaluation. For a source gait sequence, we gradually increase the number of inserted frames and run until a successful attack. Under this setting, the evaluation metric is the mean of the inserted frame numbers. We perform nontargeted attack on GaitSet and obtain results in Table~\ref{tab:number}. We can observe inserting only three adversarial frames can lead to a 96.29$\%$ fooling rate. Furthermore, for a 30 frame-length sequence, the mean frame number to 100$\%$ deceive GaitSet is only 1.6.
	
	\begin{table}
		\caption{Study on number of inserted adversarial frames.}
		\label{tab:number}
		\setlength{\tabcolsep}{2mm}
		\renewcommand{\arraystretch}{1.1}
		\begin{center}
			\begin{tabular}{|l|l|l|l|l|l|}
				\hline
				\multicolumn{6}{|c|}{dynamic evaluation}\\
				\hline
				frame number & 1     & 2   & 3   & 4 & 5\\ \hline
				fooling rate  & 52.49$\%$ & 85.27$\%$ & 96.29$\%$ & 98.43$\%$ & 99.44$\%$ \\ \hline
				\multicolumn{6}{|c|}{static evaluation}\\
				\hline
				\multicolumn{3}{|c|}{fooling rate} & \multicolumn{3}{c|}{100$\%$}\\
				\hline
				\multicolumn{3}{|c|}{mean frame number} & \multicolumn{3}{c|}{1.6}\\
				\hline
			\end{tabular}
		\end{center}
	\end{table}
	
	\section{Conclusion}
	
	In this paper, we propose a novel temporal sparse adversarial attack on gait recognition. Our method achieves good imperceptibility and a high attack success rate. Experiments on CASIA datasets indicate that the state-of-the-art model, GaitSet, is vulnerable to our adversarial attack. This reveals a key limitation in adversarial robustness research on gait recognition that requires urgent attention. Our method also shows a potential threat in practical applications as it is flexible in either attacking on biometric samples captured by a sensor or directly modifying probes. We mainly focus on the vulnerability of sequence-based models and show template features like GEI may resist our attack. The results highlight the inherent loss of temporal and fine-grained spatial information in gait templates; consequently, they can avoid deliberate attacks on vulnerable temporal features. Therefore, we identify the need for the community to consider the robustness of sequence-based methods, which possess the benefit of high accuracy, in future research. %We mainly focus on high-quality gait silhouettes, instead of RGB gait images, generation for adversarial attacks in this paper. Gait silhouette frames can be easily transferred to the source video frames with a generator like the StyleGAN~\cite{Karras2018ASG}, which is left for our future work. It is also easy to transplant our proposed approach to the targeted attack.
	
	\bibliographystyle{IEEEtran}
	\bibliography{egbib}

\begin{thebibliography}{10}
\providecommand{\url}[1]{#1}
\csname url@samestyle\endcsname
\providecommand{\newblock}{\relax}
\providecommand{\bibinfo}[2]{#2}
\providecommand{\BIBentrySTDinterwordspacing}{\spaceskip=0pt\relax}
\providecommand{\BIBentryALTinterwordstretchfactor}{4}
\providecommand{\BIBentryALTinterwordspacing}{\spaceskip=\fontdimen2\font plus
\BIBentryALTinterwordstretchfactor\fontdimen3\font minus
  \fontdimen4\font\relax}
\providecommand{\BIBforeignlanguage}[2]{{%
\expandafter\ifx\csname l@#1\endcsname\relax
\typeout{** WARNING: IEEEtran.bst: No hyphenation pattern has been}%
\typeout{** loaded for the language `#1'. Using the pattern for}%
\typeout{** the default language instead.}%
\else
\language=\csname l@#1\endcsname
\fi
#2}}
\providecommand{\BIBdecl}{\relax}
\BIBdecl

\bibitem{Wu2016ACS}
Z.~Wu, Y.~Huang, L.~Wang, X.~Wang, and T.~Tan, ``A comprehensive study on
  cross-view gait based human identification with deep cnns,'' \emph{IEEE
  Transactions on Pattern Analysis and Machine Intelligence}, vol.~39, pp.
  209--226, 2016.

\bibitem{Takemura2019OnIA}
N.~Takemura, Y.~Makihara, D.~Muramatsu, T.~Echigo, and Y.~Yagi, ``On
  input/output architectures for convolutional neural network-based cross-view
  gait recognition,'' \emph{IEEE Transactions on Circuits and Systems for Video
  Technology}, vol.~29, pp. 2708--2719, 2019.

\bibitem{He2019MultiTaskGF}
Y.~He, J.~Zhang, H.~Shan, and L.~Wang, ``Multi-task gans for view-specific
  feature learning in gait recognition,'' \emph{IEEE Transactions on
  Information Forensics and Security}, vol.~14, pp. 102--113, 2019.

\bibitem{Wolf2016MultiviewGR}
T.~Wolf, M.~Babaee, and G.~Rigoll, ``Multi-view gait recognition using 3d
  convolutional neural networks,'' in \emph{IEEE International Conference on
  Image Processing}, 2016, pp. 4165--4169.

\bibitem{Liao2017PoseBasedTN}
R.~Liao, C.~Cao, E.~B. Garcia, S.~Yu, and Y.~Huang, ``Pose-based
  temporal-spatial network (ptsn) for gait recognition with carrying and
  clothing variations,'' in \emph{Chinese Conference on Biometric Recognition},
  2017, pp. 474--483.

\bibitem{Jia2019AttackingGR}
M.~Jia, H.~Yang, D.~Huang, and Y.~Wang, ``Attacking gait recognition systems
  via silhouette guided gans,'' in \emph{Proceedings of the 27th ACM
  International Conference on Multimedia}, 2019, pp. 638--646.

\bibitem{DBLP1}
D.~Gafurov, E.~Snekkenes, and P.~Bours, ``Spoof attacks on gait authentication
  system,'' \emph{{IEEE} Trans. Information Forensics and Security}, pp.
  491--502, 2007.

\bibitem{DBLP2}
A.~Hadid, M.~Ghahramani, V.~Kellokumpu, M.~Pietik{\"{a}}inen, J.~D. Bustard,
  and M.~S. Nixon, ``Can gait biometrics be spoofed?'' in \emph{International
  Conference on Pattern Recognition}, 2012, pp. 3280--3283.

\bibitem{DBLP3}
A.~Hadid, M.~Ghahramani, V.~Kellokumpu, X.~Feng, J.~D. Bustard, and M.~S.
  Nixon, ``Gait biometrics under spoofing attacks: an experimental
  investigation,'' \emph{J. Electronic Imaging}, vol.~24, no.~6, p. 063022,
  2015.

\bibitem{Szegedy2013IntriguingPO}
C.~Szegedy, W.~Zaremba, I.~Sutskever, J.~Bruna, D.~Erhan, I.~J. Goodfellow, and
  R.~Fergus, ``Intriguing properties of neural networks,'' in
  \emph{International Conference on Learning Representations}, 2014.

\bibitem{Goodfellow2014ExplainingAH}
I.~J. Goodfellow, J.~Shlens, and C.~Szegedy, ``Explaining and harnessing
  adversarial examples,'' in \emph{International Conference on Learning
  Representations}, 2015.

\bibitem{Xie2017AdversarialEF}
C.~Xie, J.~Wang, Z.~Zhang, Y.~Zhou, L.~Xie, and A.~L. Yuille, ``Adversarial
  examples for semantic segmentation and object detection,'' in \emph{IEEE
  International Conference on Computer Vision}, 2017, pp. 1378--1387.

\bibitem{Sharif2016AccessorizeTA}
M.~Sharif, S.~Bhagavatula, L.~Bauer, and M.~K. Reiter, ``Accessorize to a
  crime: Real and stealthy attacks on state-of-the-art face recognition,'' in
  \emph{Proceedings of the 2016 ACM SIGSAC Conference on Computer and
  Communications Security}, 2016, pp. 1528--1540.

\bibitem{Madry2017TowardsDL}
A.~Madry, A.~Makelov, L.~Schmidt, D.~Tsipras, and A.~Vladu, ``Towards deep
  learning models resistant to adversarial attacks,'' \emph{ArXiv}, vol.
  abs/1706.06083, 2017.

\bibitem{Chao2018GaitSetRG}
H.~Chao, Y.~He, J.~Zhang, and J.~Feng, ``Gaitset: Regarding gait as a set for
  cross-view gait recognition,'' in \emph{Proceedings of the AAAI Conference on
  Artificial Intelligence}, 2019, pp. 8126--8133.

\bibitem{Brown2018UnrestrictedAE}
T.~B. Brown, N.~Carlini, C.~Zhang, C.~Olsson, P.~F. Christiano, and I.~J.
  Goodfellow, ``Unrestricted adversarial examples,'' \emph{ArXiv}, vol.
  abs/1809.08352, 2018.

\bibitem{DBLP:conf/icb/ShiragaMMEY16}
K.~Shiraga, Y.~Makihara, D.~Muramatsu, T.~Echigo, and Y.~Yagi, ``Geinet:
  View-invariant gait recognition using a convolutional neural network,'' in
  \emph{International Conference on Biometrics}, 2016, pp. 1--8.

\bibitem{DBLP:conf/cvpr/YuCRP17}
S.~Yu, H.~Chen, E.~B.~G. Reyes, and N.~Poh, ``Gaitgan: Invariant gait feature
  extraction using generative adversarial networks,'' in \emph{IEEE Conference
  on Computer Vision and Pattern Recognition Workshops}, 2017, pp. 532--539.

\bibitem{Han2006IndividualRU}
J.~Han and B.~Bhanu, ``Individual recognition using gait energy image,''
  \emph{IEEE Transactions on Pattern Analysis and Machine Intelligence},
  vol.~28, pp. 316--322, 2006.

\bibitem{Yu2006AFF}
S.~Yu, D.~Tan, and T.~Tan, ``A framework for evaluating the effect of view
  angle, clothing and carrying condition on gait recognition,'' in
  \emph{International Conference on Pattern Recognition}, 2006, pp. 441--444.

\bibitem{Kurakin2016AdversarialML}
A.~Kurakin, I.~J. Goodfellow, and S.~Bengio, ``Adversarial machine learning at
  scale,'' \emph{ArXiv}, vol. abs/1611.01236, 2016.

\bibitem{MoosaviDezfooli2015DeepFoolAS}
S.-M. Moosavi-Dezfooli, A.~Fawzi, and P.~Frossard, ``Deepfool: A simple and
  accurate method to fool deep neural networks,'' in \emph{IEEE Conference on
  Computer Vision and Pattern Recognition}, 2016, pp. 2574--2582.

\bibitem{Papernot2015TheLO}
N.~Papernot, P.~D. McDaniel, S.~Jha, M.~Fredrikson, Z.~B. Celik, and A.~Swami,
  ``The limitations of deep learning in adversarial settings,'' in \emph{IEEE
  European Symposium on Security and Privacy}, 2016, pp. 372--387.

\bibitem{Dong2017BoostingAA}
Y.~Dong, F.~Liao, T.~Pang, H.~Su, J.~Zhu, X.~Hu, and J.~Li, ``Boosting
  adversarial attacks with momentum,'' in \emph{IEEE Conference on Computer
  Vision and Pattern Recognition}, 2018, pp. 9185--9193.

\bibitem{Song2018ConstructingUA}
Y.~Song, R.~Shu, N.~Kushman, and S.~Ermon, ``Constructing unrestricted
  adversarial examples with generative models,'' in \emph{Advances in Neural
  Information Processing Systems}, 2018, pp. 8312--8323.

\bibitem{Poursaeed2019FinegrainedSO}
O.~Poursaeed, T.~Jiang, H.~Yang, S.~J. Belongie, and S.-N. Lim, ``Fine-grained
  synthesis of unrestricted adversarial examples,'' \emph{ArXiv}, vol.
  abs/1911.09058, 2019.

\bibitem{Karras2018ASG}
T.~Karras, S.~Laine, and T.~Aila, ``A style-based generator architecture for
  generative adversarial networks,'' in \emph{IEEE Conference on Computer
  Vision and Pattern Recognition}, 2019, pp. 4401--4410.

\bibitem{Wei2018SparseAP}
X.~Wei, J.~Zhu, and H.~Su, ``Sparse adversarial perturbations for videos,'' in
  \emph{Proceedings of the AAAI Conference on Artificial Intelligence}, 2018,
  pp. 8973--8980.

\bibitem{Chen2019AppendingAF}
Z.~Chen, L.~Xie, S.~Pang, Y.~He, and Q.~Tian, ``Appending adversarial frames
  for universal video attack,'' \emph{ArXiv}, vol. abs/1912.04538, 2019.

\bibitem{Goodfellow2014GenerativeAN}
I.~J. Goodfellow, J.~Pouget-Abadie, M.~Mirza, B.~Xu, D.~Warde-Farley, S.~Ozair,
  A.~C. Courville, and Y.~Bengio, ``Generative adversarial nets,'' in
  \emph{Advances in Neural Information Processing Systems}, 2014, pp.
  2672--2680.

\bibitem{DBLP:journals/corr/ArjovskyCB17}
M.~Arjovsky, S.~Chintala, and L.~Bottou, ``Wasserstein {GAN},'' \emph{Arxiv},
  vol. abs/1701.07875, 2017.

\bibitem{DBLP:conf/nips/GulrajaniAADC17}
I.~Gulrajani, F.~Ahmed, M.~Arjovsky, V.~Dumoulin, and A.~C. Courville,
  ``Improved training of wasserstein gans,'' in \emph{Advances in Neural
  Information Processing Systems}, 2017, pp. 5767--5777.

\bibitem{DBLP:conf/cvpr/IsolaZZE17}
P.~Isola, J.~Zhu, T.~Zhou, and A.~A. Efros, ``Image-to-image translation with
  conditional adversarial networks,'' in \emph{IEEE Conference on Computer
  Vision and Pattern Recognition}, 2017, pp. 5967--5976.

\bibitem{DBLP:conf/cvpr/Park0WZ19}
T.~Park, M.~Liu, T.~Wang, and J.~Zhu, ``Semantic image synthesis with
  spatially-adaptive normalization,'' in \emph{IEEE Conference on Computer
  Vision and Pattern Recognition}, 2019, pp. 2337--2346.

\bibitem{PULSE}
S.~Menon, A.~Damian, S.~Hu, N.~Ravi, and C.~Rudin, ``{PULSE:} self-supervised
  photo upsampling via latent space exploration of generative models,'' in
  \emph{IEEE Conference on Computer Vision and Pattern Recognition}, 2020, pp.
  2434--2442.

\bibitem{Papernot2016TransferabilityIM}
N.~Papernot, P.~D. McDaniel, and I.~J. Goodfellow, ``Transferability in machine
  learning: from phenomena to black-box attacks using adversarial samples,''
  \emph{arXiv preprint arXiv:1605.07277}, 2016.

\bibitem{Liu2016DelvingIT}
Y.~Liu, X.~Chen, C.~Liu, and D.~X. Song, ``Delving into transferable
  adversarial examples and black-box attacks,'' in \emph{Proceedings of
  International Conference on Learning Representations}, 2017.

\bibitem{liu2021selfgait}
Y.~Liu, Y.~Zeng, J.~Pu, H.~Shan, P.~He, and J.~Zhang, ``Selfgait: A
  spatiotemporal representation learning method for self-supervised gait
  recognition,'' in \emph{IEEE International Conference on Acoustics, Speech
  and Signal Processing}, 2021, pp. 2570--2574.

\end{thebibliography}

\end{document}